\documentclass[10pt]{article} % For LaTeX2e
\usepackage[preprint]{styles/tmlr}

\usepackage{amsmath,amsfonts,bm}

\def\eqref#1{equation~\ref{#1}}
\def\1{\bm{1}}

\def\vs{{\bm{s}}}

\DeclareMathAlphabet{\mathsfit}{\encodingdefault}{\sfdefault}{m}{sl}
\SetMathAlphabet{\mathsfit}{bold}{\encodingdefault}{\sfdefault}{bx}{n}

\DeclareMathOperator*{\argmin}{arg\,min}

\usepackage{hyperref}
\usepackage{url}

\usepackage{subcaption}  % FOR subfigures
\usepackage{graphicx}
\usepackage{booktabs}
\usepackage{multirow}
\usepackage{microtype} % For organizing space (allowed)

\usepackage{algorithm}
\usepackage{algpseudocode}

\usepackage[capitalize]{cleveref}

\newtheorem{theorem}{Theorem}
\newtheorem{proposition}{Proposition}
\newtheorem{definition}{Definition}

\DeclareMathOperator{\similarity}{sim}

\RequirePackage{xspace}
\makeatletter
\DeclareRobustCommand\onedot{\futurelet\@let@token\@onedot}
\def\@onedot{\ifx\@let@token.\else.\null\fi\xspace}

\def\eg{\emph{e.g}\onedot} 

\def\ie{\emph{i.e}\onedot}

\def\vs{\emph{vs}\onedot}
\def\wrt{w.r.t\onedot} 
\makeatother

\usepackage[dvipsnames]{xcolor}
\usepackage{soul}

\title{A Mutual Information Perspective on Multiple Latent \\ Variable Generative Models for Positive View Generation}

\author{\name Dario Serez \email dario.serez@iit.it \\
      \addr Istituto Italiano di Tecnologia, Italy
      \AND
      \name Marco Cristani \email marco.cristani@univr.it \\
      \addr Reykjavík University, Iceland \\
            University of Verona, Italy
      \AND
      \name Alessio Del Bue \email alessio.delbue@iit.it \\
      \addr Istituto Italiano di Tecnologia, Italy
      \AND
      \name Vittorio Murino \email vittorio.murino@iit.it \\
      \addr Istituto Italiano di Tecnologia, Italy \\
            University of Verona, Italy
      \AND
      \name Pietro Morerio \email pietro.morerio@iit.it \\
      \addr Istituto Italiano di Tecnologia, Italy}

\def\month{09}  % Insert correct month for camera-ready version
\def\year{2025} % Insert correct year for camera-ready version
\def\openreview{\url{https://openreview.net/forum?id=uaj8ZL2PtK}} % Insert correct link to OpenReview for camera-ready version

\begin{document}

\maketitle

\begin{abstract}
In image generation, Multiple Latent Variable Generative Models (MLVGMs) employ multiple latent variables to gradually shape the final images, from global characteristics to finer and local details (\eg, StyleGAN, NVAE), emerging as powerful tools for diverse applications. 
Yet their generative dynamics remain only empirically observed, without a systematic understanding of each latent variable's impact.
%and latent variable utilization remain only empirically observed.
In this work, we propose a novel framework that quantifies the contribution of each latent variable using Mutual Information (MI) as a metric. Our analysis reveals that current MLVGMs often underutilize some latent variables, and provides actionable insights for their use in downstream applications.
%to systematically quantify the impact of each latent variable in MLVGMs, using Mutual Information (MI) as a guiding metric. Our analysis reveals underutilized variables and can guide the use of MLVGMs in downstream applications.

With this foundation, we introduce a method for generating synthetic data for Self-Supervised Contrastive Representation Learning (SSCRL). By leveraging the hierarchical and disentangled variables of MLVGMs, our approach produces diverse and semantically meaningful views without the need for real image data. 
%and guided by the previous analysis, we apply tailored latent perturbations to produce diverse views for SSCRL, without relying on real data altogether.
Additionally, we introduce a Continuous Sampling (CS) strategy, where the generator dynamically creates new samples during SSCRL training, greatly increasing data variability. Our comprehensive experiments demonstrate the effectiveness of these contributions, showing that MLVGMs' generated views compete on par with or even surpass views generated from real data. 

This work establishes a principled approach to understanding and exploiting MLVGMs, advancing both generative modeling and self-supervised learning. Code and pre-trained models at: \hyperlink{https://github.com/SerezD/mi_ml_gen}{https://github.com/SerezD/mi\_ml\_gen}.

\end{abstract}

\begin{figure}[ht]
    \begin{center}
        \begin{subfigure}{0.48\linewidth}
            \includegraphics[width=0.98\linewidth]{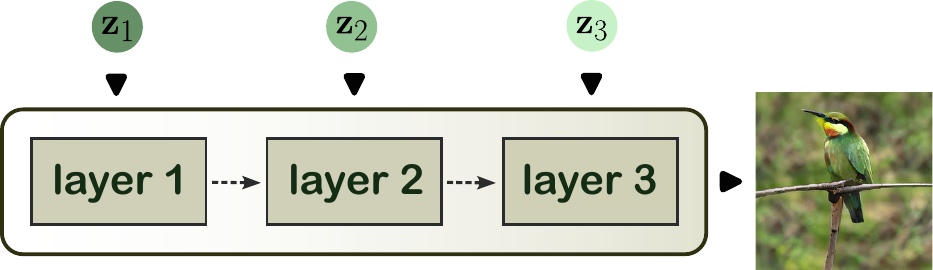}
            \caption{}
            \label{fig:1a}
        \end{subfigure}
        \hfill
        \begin{subfigure}{0.48\linewidth}
            \includegraphics[width=1.00\linewidth]{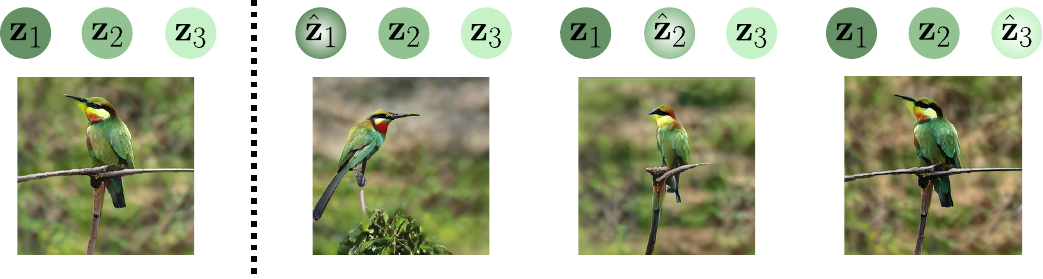}
            \caption{}
            \label{fig:1b}
        \end{subfigure}
    \end{center}

  \caption{\textbf{(a)} Multiple Latent Variable Generative Models utilize multiple latent variables (here $\mathbf{Z}_1, \mathbf{Z}_2, \mathbf{Z}_3$), which are sequentially introduced at different layers of the generative network to produce high-quality images. \textbf{(b)} The base image (left), generated using points $\mathbf{z}_1, \mathbf{z}_2, \mathbf{z}_3$, can be selectively modified by altering individual latents ($\mathbf{z}_1$ to $\hat{\mathbf{z}}_1$, $\mathbf{z}_2$ to $\hat{\mathbf{z}}_2$, or $\mathbf{z}_3$ to $\hat{\mathbf{z}}_3$). Each latent affects the final image differently, at first influencing broader, global attributes and later refining finer, local details (darker to lighter shading in the figure).}
  
  \label{fig:1}
\end{figure}

\section{Introduction}
\label{sec:introduction}

% 1. WHAT IS THE PROBLEM? 
\textit{Latent Variable Generative Models} (LVGMs), such as %including
Variational Autoencoders (VAEs) \citep{VAE-1,VAE-2} and Generative Adversarial Networks (GANs) \citep{GAN}, are foundational approaches for image generation. Given a random variable $\mathbf{X} \in \mathcal{X}$, representing high-dimensional pictures in pixel space, LVGMs aim to approximate the underlying data distribution $p(\mathbf{X})$. To achieve this, they learn a parameterized generator $g(\mathbf{z}; \theta) = \mathbf{x}$, where $\mathbf{Z} \in \mathcal{Z}$ denotes a latent variable sampled from a simpler and known distribution in the $\mathcal{Z}$ latent space. A key objective of the learning process is to ensure that the generator is continuous, such that neighboring latent points $\mathbf{z}'$ and $\mathbf{z}''$ are mapped to perceptually similar outputs $\mathbf{x}'$ and $\mathbf{x}''$. This regularization of the latent space allows LVGMs to generate novel content and meaningfully interpolate latent features \citep{dcgan,betavae}.

Over the years, advancements in latent generative modeling have focused on the use of \emph{multiple} latent variables, rather than a single latent code \citep{nvae,style-gan,style-gan2,style-gan3,Style-GAN-XL}. By incorporating latent variables at different layers of the network (see \Cref{fig:1a}), these Multiple Latent Variable Generative Models (MLVGMs) offer a hierarchical structure where early latent codes influence broad, global features and later codes refine finer, local details (\Cref{fig:1b}). The resulting architecture enhances control over the image synthesis process, enabling the generation of high-resolution images with richer detail and improved precision.
%Over the years, significant advancements have been made in latent generative modeling architectures \citep{nvae,style-gan,style-gan2,style-gan3,Style-GAN-XL}, particularly through the introduction of \emph{multiple} latent variables that are progressively integrated into the network (see \Cref{fig:1a}). This hierarchical design enhances control over the generation process, as different latent variables influence distinct aspects of the output image, ranging from coarse-grained global characteristics to finer, localized features (\Cref{fig:1b}). Consequently, modern \emph{Multiple Latent Variable Generative Models} (MLVGMs) can generate high-resolution images with improved precision and richer detail.

% 2. WHY IS IT IMPORTANT/WHAT HAVE OTHER PEOPLE DONE?
The ability to disentangle global and local features in image generation not only improves the visual quality of generated images but also widens the application scope of these models.
%One major benefit of this seemingly simple improvement is the potential application of MLVGMs across various tasks. 
%Notably, 
For instance, the StyleGAN architecture \citep{style-gan} has demonstrated exceptional performance in image editing \citep{stylegan_edit1,stylegan_edit2}, manipulation \citep{stylegan_manipulation}, and translation \citep{stylegan_translation}. Additionally, recent studies have shown that MLVGMs can serve as effective foundation models for tasks such as adversarial purification \citep{serez2024mlvgms}. Collectively, these findings highlight the versatility of MLVGMs, showcasing their utility not only in creative and generative domains but also as pre-trained models for broader applications.

% Why previous work is not sufficient? What do we propose to do differently?
Nevertheless, existing research primarily leverages the ``global-to-local'' behavior of MLVGMs as an empirical tool, applying it across diverse tasks without delving into the mechanics of latent variable utilization. 
In other terms, these approaches assume that earlier latent variables shape coarse image attributes while later ones refine fine details, but they do so without formally analyzing how each latent variable contributes to image generation. As a result, the internal dynamics of MLVGMs remain poorly understood.
%Despite their effectiveness, these methods fail to investigate how each latent variable, introduced at different stages of the generative process, contributes to the hierarchical refinement of global and local image features. 

\begin{figure}[t]
    \centering
    \includegraphics[width=0.85\linewidth]{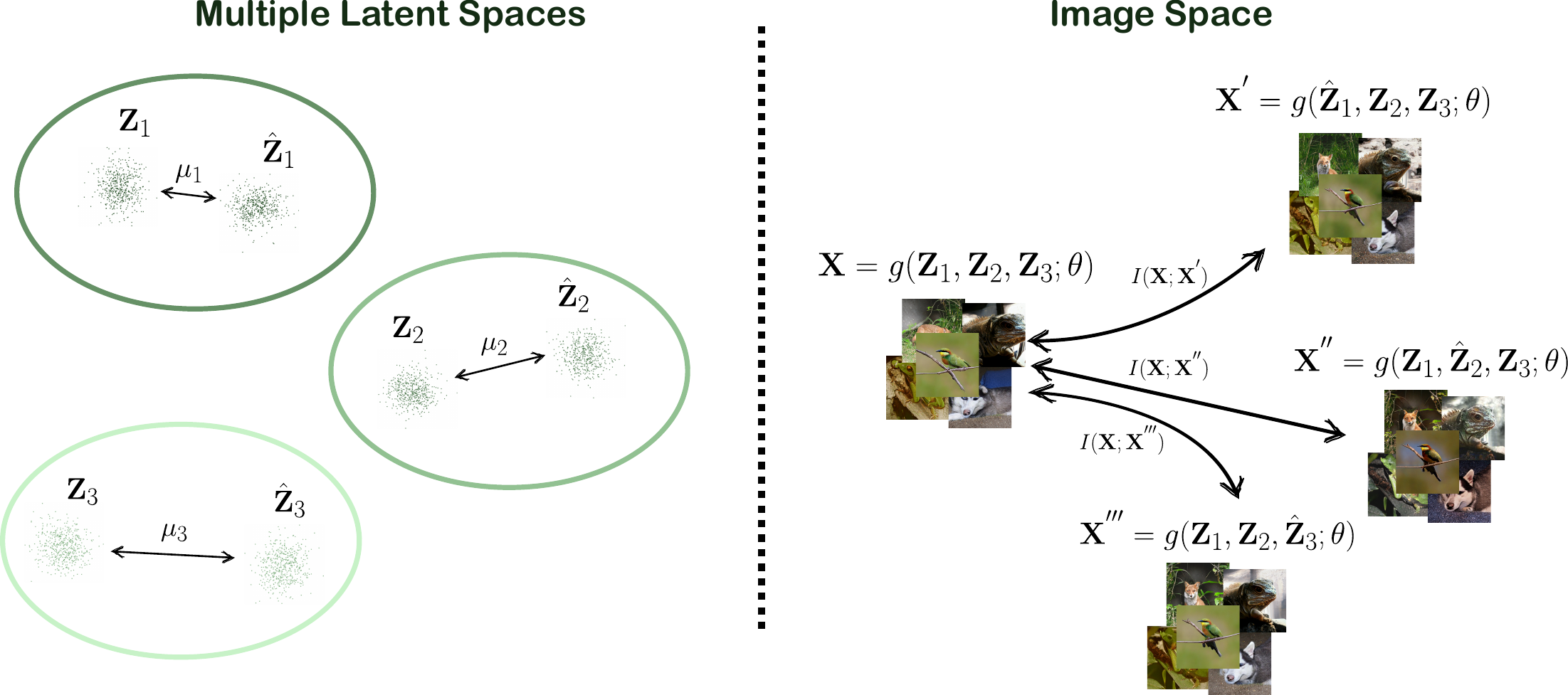}
  \caption{Illustration of our findings on the ``global-to-local'' behavior in MLVGMs. On the left, perturbations are applied to each latent variable independently ($\mathbf{Z}_1 \to \hat{\mathbf{Z}}_1$, $\mathbf{Z}_2 \to \hat{\mathbf{Z}}_2$, and $\mathbf{Z}_3 \to \hat{\mathbf{Z}}_3$), with the average perturbation magnitude increasing across latent spaces ($\mu_1 < \mu_2 < \mu_3$). On the right, each perturbed latent variable is used to generate modified images ($\mathbf{X}'$, $\mathbf{X}''$, and $\mathbf{X}'''$) from the original $\mathbf{X}$. Notably, the increasing perturbation magnitude in the latent space maintains approximately equal Mutual Information shifts in the image space: $I(\mathbf{X}; \mathbf{X}') \approx I(\mathbf{X}; \mathbf{X}'') \approx I(\mathbf{X}; \mathbf{X}''')$. This provides the first quantitative measure of the ``global-to-local'' property, where earlier latents affect global features and later latents refine local details.}
  \label{fig:2}
\end{figure}

To address this gap, we propose a novel approach that establishes a direct relationship between feature distances in each latent space ($\mathbf{Z}_1, \mathbf{Z}_2, \dots, \mathbf{Z}_n$) and mutual information (MI) shifts in the shared image space $\mathbf{X}$. 
The key insight is that producing an equivalent MI shift in the output space requires increasingly larger perturbations ($\mu_i$) as we move deeper into the generative hierarchy—i.e., from $\mathbf{Z}_1$ to $\mathbf{Z}_n$. This quantitatively confirms the global-to-local pattern and reveals how influence diminishes across successive latent variables (see \Cref{fig:2}).
%Our analysis reveals that \textit{achieving equivalent MI shifts by varying a single latent variable} (\eg $\mathbf{Z}_i$ to $\hat{\mathbf{Z}}_i$) \textit{requires progressively larger average perturbations} ($\mu_i$ - see section \ref{subsec:mlvgms}) as the $i$-th variable is introduced later in the generative process. This observation aligns with the intuition that the influence of individual latent variables diminishes throughout the generative hierarchy (see \Cref{fig:2}).

% Why is it better?
By grounding this analysis in information theory, our approach moves beyond intuition and provides a principled framework for understanding latent variable roles in MLVGMs. It enables direct comparisons between latent spaces and exposes inefficiencies in how models allocate representational power. Notably, we reveal the underutilization of later-stage latents in all tested models, opening to further architectural improvements and enabling better strategies for leveraging MLVGMs in downstream tasks.
%Our systematic evaluation offers a quantitative framework for understanding the contribution of each latent variable in MLVGMs, extending beyond empirical observations. Notably, we also find that the latter latent variables are often underutilized in modern MLVGMs, highlighting potential inefficiencies in current training paradigms. With these insights, we not only provide a deeper understanding of MLVGMs but also can guide their effective utilization in downstream tasks.

% Explain the view generation method
With
%Building on 
this understanding, we propose a novel application of MLVGMs in Self-Supervised Contrastive Representation Learning (SSCRL). In SSCRL, feature extractors, or encoders $f(\mathbf{x}; \phi)$ with parameters $\phi$, are trained to represent data by contrasting positive and negative views. Positive views are semantically similar images, encouraged to have close representations in the latent space, while negative views correspond to unrelated data points that are forced to have distant representations. 
Therefore, we propose to leverage the different impacts of multiple latent variables in MLVGMs to manipulate specific features and generate positive views. This approach enables the training of SSCRL encoders \emph{without relying on real data}, demonstrating the potential of MLVGMs as pre-trained models for producing high-quality synthetic images tailored for representation learning.

\begin{figure}[tb]
    \begin{center}
        \begin{subfigure}{0.46\linewidth}
            \includegraphics[width=1.0\linewidth]{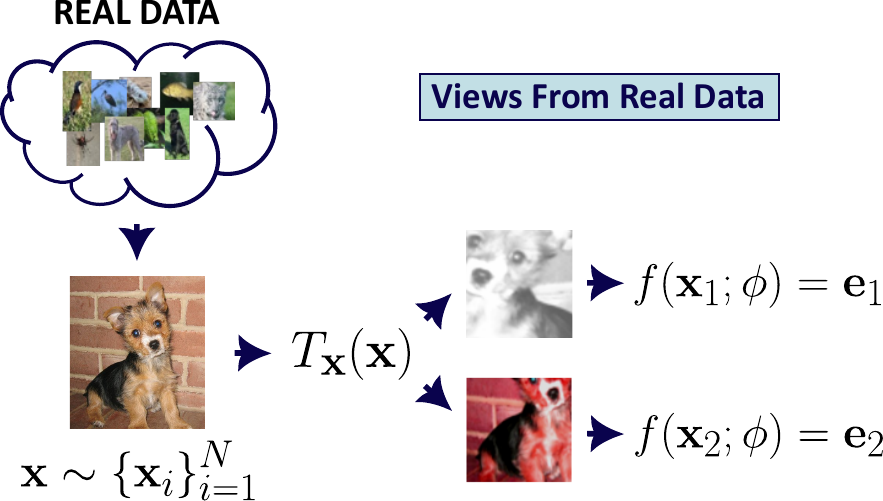}
            \caption{}
            \label{fig:3a}
        \end{subfigure}
        \hfill
        \begin{subfigure}{0.43\linewidth}
            \includegraphics[width=1.0\linewidth]{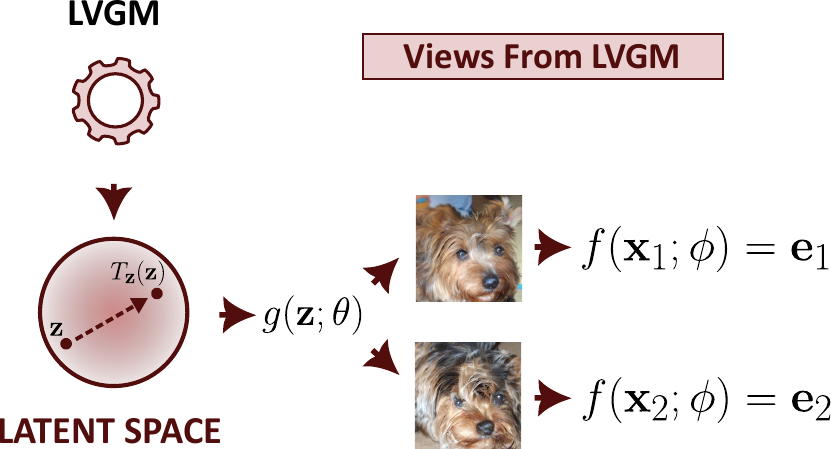}
            \caption{}
            \label{fig:3b}
        \end{subfigure}
        \hfill
        \vspace{12pt}
        \begin{subfigure}{0.75\linewidth}
            \includegraphics[width=1.0\linewidth]{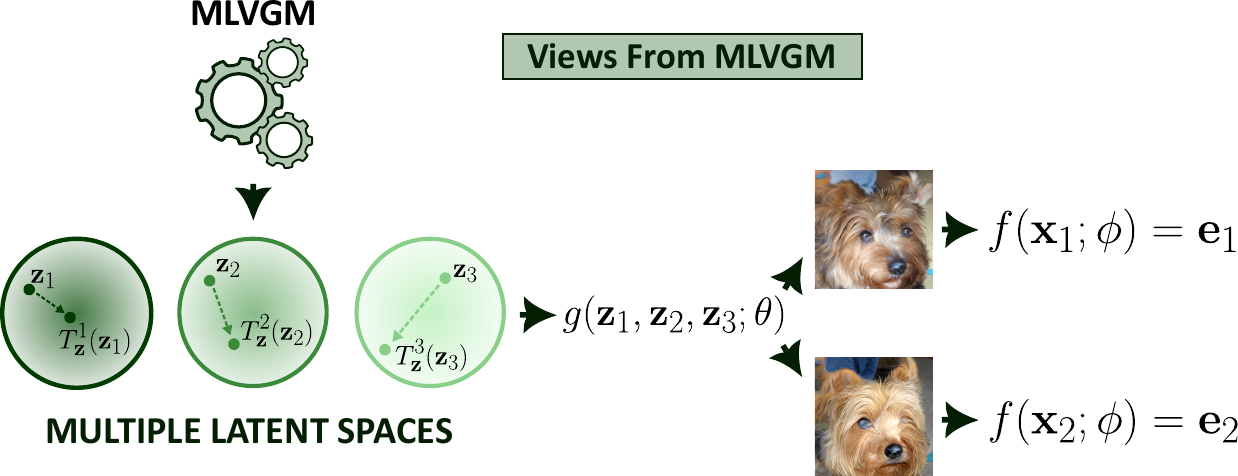}
            \caption{}
            \label{fig:3c}
        \end{subfigure}
    \end{center}

  \caption{Self-Supervised Contrastive Representation Learning (SSCRL) optimizes an embedding function $f(\mathbf{x}; \phi) = \mathbf{e}$, mapping semantically similar images $\mathbf{x}_1, \mathbf{x}_2$ to nearby latent representations $\mathbf{e}_1, \mathbf{e}_2$. \textbf{(a)} In the classic approach, positive views are generated by applying hand-crafted transformations in the pixel space, $T_{\mathbf{x}}$, to a finite dataset of images $\mathbf{x} = \{\mathbf{x}_i\}_{i=1}^N$. \textbf{(b)} Alternatively, positive views can be generated by sampling nearby points in the latent space of a Latent Variable Generative Model (LVGM), $g(\mathbf{z}; \theta)$. However, since image features remain highly entangled in the latent space, even subtle perturbations $T_{\mathbf{z}}(\mathbf{z})$ may change important characteristics such as dog breed (Australian to Yorkshire Terrier in the figure). \textbf{(c)} Our framework leverages a Multiple Latent Variable Generative Model (MLVGM), represented as $g(\mathbf{z}_1, \mathbf{z}_2, \dots, \mathbf{z}_n; \theta)$ ($n=3$ in the figure). By applying tailored perturbations $T_{\mathbf{z}}^i(\mathbf{z}_i)$ to each latent variable, we leverage the hierarchical feature representation to obtain a broader range of valid transformations while maintaining important semantic aspects.}
  \label{fig:3}
\end{figure}

The primary objective of SSCRL is to enforce a desired set of invariances in the learned representations \citep{contrastive-invariant}, achieved by creating valid positive views. \Cref{fig:3} compares the proposed method with standard pixel-space augmentations and single latent variable generative models (LVGMs) for view generation. 
In the typical approach (\Cref{fig:3a}), a finite set of hand-crafted transformations, such as color adjustments, cropping, or flipping, is applied directly in the pixel space. Alternatively, invariances can be introduced at the latent level of a pre-trained LVGM (\Cref{fig:3b}). However, in LVGMs, all image features are entangled within a single latent space, making it difficult to generate specific invariances (\eg, altering fur patterns) without inadvertently affecting global features, such as changing the dog breed (\eg, from Australian Terrier to Yorkshire Terrier in the figure). 
In contrast, MLVGMs inherently disentangle global and local features, enabling precise control over specific characteristics. For instance, using MLVGMs, attributes like fur patterns or color can be modified while preserving global features, such as the dog breed. This is achieved by independently perturbing each latent variable to a desired magnitude, as illustrated in \Cref{fig:3c}.

% Here explain also the problem of real data vs synthetic data (Continuous Sampling). 
The use of generative models to create both anchor and positive views introduces a significant challenge: the lower classification accuracy typically observed when training on synthetic data compared to real data \citep{cas}. Prior studies, such as \citet{supclassifier3,supclassifier7}, have identified the lack of diversity in generated images as a primary factor contributing to this issue. To mitigate this, these works propose increasing dataset diversity by sampling and storing a large number of synthetic images before training. 
In contrast, we propose a novel approach called \emph{Continuous Sampling} (CS) to address this limitation. With CS, new images are generated ``online'' during the encoder network's training process, offering three key advantages: 
(i) no need to store large quantities of synthetic data, thereby preserving disk space; (ii) no data loading step, which is often the bottleneck in neural network training, as new batches are generated directly into GPU memory; and (iii) maximized diversity, specifically by ensuring that each batch is freshly sampled and never reused, unlike prior methods that rely on a fixed-size synthetic dataset. 

% Some details on experiments
To evaluate our contributions, we apply our novel quantification algorithm to two distinct MLVGMs: a BigBiGAN \citep{bigbigan} pre-trained on ImageNet-$1$K \citep{imagenet}, and a StyleGAN2 \citep{style-gan2} pre-trained on LSUN Cars \citep{lsun}. Subsequently, we leverage the same MLVGMs to generate views for SSCRL using our proposed Continuous Sampling (CS) strategy. Specifically, we train feature extractors with three different SSCRL frameworks, namely SimCLR \citep{simclr}, SimSiam \citep{simsiam}, and BYOL \citep{byol}. The learned representations are then validated following a consolidated practice through linear classification across multiple downstream datasets and object detection on Pascal VOC \citep{pascal-voc}. Our results demonstrate that MLVGM-based view generation outperforms simple LVGM-based techniques and \textit{achieves comparable or superior results than training with real data}. Additionally, we measure the training time per epoch using Continuous Sampling \vs standard data loading, establishing CS as an efficient alternative for increasing data diversity.

% sum up contributions
To sum up, our contributions are as follows: 
i) We propose the first method to quantify the influence of individual latent variables in Multiple Latent Variable Generative Models (MLVGMs), which can reveal underutilized latent spaces and serve as a helpful tool for downstream applications.
ii) We leverage the natural disentanglement of coarse from fine features in MLVGMs to create positive views for Self-Supervised Contrastive Representation Learning (SSCRL), enabling tailored invariances that outperform previous methods using both real and synthetic data.
iii) We introduce Continuous Sampling, a novel procedure that dynamically generates new batches during SSCRL training, increasing data diversity, reducing storage requirements, and maintaining competitive training time performance.

\section{Related Works}
\label{sec:rel_works}

\paragraph{MLVGMs.}
The idea of utilizing multiple latent variables is well-established in the generative models' literature, typically presented as an evolution of Latent Variable Generative Models (LVGMs). For instance, Variational Autoencoders (VAEs) \citep{VAE-1,VAE-2} leverage multiple latent variables to enhance the expressivity of approximate distributions, as demonstrated by architectures such as NVAE \citep{nvae} and VD-VAE \citep{vdvae}, or to improve latent disentanglement, as in \citet{disentangled-vlae}. Similarly, Generative Adversarial Networks (GANs) \citep{GAN} have embraced this concept in models like LapGAN \citep{lap-gan}, BigGANs \citep{biggan,bigbigan}, and GigaGAN \citep{Giga-GAN}. Advances in Normalizing Flows \citep{NF-1,NF-2} have also incorporated multiple latent variables, with works like \citet{rg-flow} introducing architectures inspired by physics to achieve this goal.

In such a growing %rich 
environment, numerous applications of MLVGMs have emerged. Of particular relevance in this context is the StyleGAN family \citep{style-gan,style-gan2,style-gan3,Style-GAN-XL}, which has been widely applied in image editing and manipulation tasks \citep{stylegan_manipulation,stylegan_translation,stylegan_edit1,stylegan_edit2}.  More recently, MLVGMs have also been used as foundation models for non-generative downstream tasks, such as purification against adversarial attacks \citep{serez2024mlvgms}. Motivated by such a rapid expansion, in this work we propose a novel application in SSCRL and for the first time study how the ``generative load'' is spread across latent variables in the current architectures.
%Motivated by this growing body of research, in this work, we address the problem of quantitatively evaluating the impact of individual latent variables in MLVGMs. To this end, we propose a novel method based on information theory, capable of revealing underutilized variables and serving as a tool for further applications. Furthermore, we employ MLVGMs in an unprecedented way, that is to generate positive views for Self-Supervised Contrastive Representation Learning (SSCRL).

\paragraph{SSCRL view generation.}

Self-Supervised Contrastive Representation Learning (SSCRL) \citep{cl-first} aims to learn meaningful latent representations without relying on labeled data, primarily by designing informative positive views \citep{infomin,contrastive-invariant}. Early approaches, such as \citet{DIM,pirl,swav}, focused on pretext tasks like matching global and local parts of an image to create multiple views. Subsequently, SimCLR \citep{simclr}, a foundational method in the field, introduced the use of manually designed transformations, including flipping, cropping, and color distortions. More recent works have explored advanced techniques, such as learning views in an adversarial manner \citep{viewmaker,adversarial} or projecting anchor images into the latent spaces of pre-trained generators \citep{sample-icgan-1,sample-icgan-2,sample-from-real-1,sample-from-real-2,sample-from-real-3,sample-from-real-4}. While our approach differs by relying solely on synthetic data, we show that it is also complementary to previous techniques, allowing to apply different transformations (denoted as $T_\mathbf{x}(\mathbf{x})$ in \Cref{fig:3a}) on top of the generated views-\ie directly in the pixel space.

%The unifying feature of these methods is their reliance on real datasets, using anchor images as the starting point. In contrast, our approach generates views exclusively from synthetic data using MLVGMs. As a consequence, our method is complementary to existing techniques, as transformations like those introduced by SimCLR (denoted as $T_\mathbf{x}(\mathbf{x})$ in \Cref{fig:3a}) can be seamlessly applied to the views generated by MLVGMs, potentially leading to even more informative representations. We test this hypothesis in the experimental section, combining pixel-space augmentations with our latent space views. 

Further along our line of work, methods like \citet{gen-rep,cop-gen} have proposed generating fully synthetic views by sampling nearby points in the latent space of pre-trained LVGMs (\Cref{fig:3b}). 
We direct compare against these baselines in the experimental section, showing that the coarse-to-fine feature disentanglement of MLVGMs allows to obtain better views for most downstream tasks.
%However, the primary limitation of these methods lies in the entanglement of all image features within a single latent space, which complicates the task of generating valid positive views. By leveraging the multiple latent spaces of MLVGMs, our approach disentangles coarse, global information from finer, local details, greatly simplifying the definition of valid views and improving the quality of the learned representations (\Cref{fig:3c}). 

Finally, recent efforts have explored the generation of synthetic views in a text-to-image setting \citep{tti-views2,tti-views1}. While this direction holds promise, particularly when combined with MLVGMs, its application to our framework remains limited. This is primarily due to the lack of publicly available code and pre-trained models for text-to-image MLVGMs, such as GigaGAN \citep{Giga-GAN}.

\paragraph{Training with generated data.}
The remarkable performance of modern generative models, such as \citet{stable-diffusion,muse}, has opened up new possibilities for using synthetic data to train classifier networks. A common strategy involves augmenting real datasets with generated samples, which has shown promise in enhancing classification performance \citep{supclassifier6,supclassifier1,supclassifier5}. Alternatively, more ambitious efforts attempt to train classifiers entirely on synthetic data, leveraging advanced text-to-image models to obtain high-quality datasets \citep{supclassifier2,supclassifier4}. 

The primary challenge in these approaches is the limited diversity of generated data, which has been identified as a key factor contributing to the performance gap between classifiers trained on real versus synthetic datasets \citep{cas}. Recent studies \citep{scaling} suggest that scaling up the size of synthetic training sets can reduce this accuracy gap, though it does not fully eliminate it. However, generating large datasets introduces its own set of challenges, particularly increased disk space usage and data management overhead. Existing methods \citep{supclassifier3,supclassifier7} address this issue by partially renewing synthetic data at each epoch or by regenerating the dataset entirely every $N$ epochs. In this work, we surpass the common assumption about the inefficiencies of generating training images in real time, leveraging fast-sampling models, such as GANs, to generate data directly during training and avoiding storage and loading bottlenecks altogether.
%In contrast, we leverage fast-sampling models, such as GANs, to generate data directly during training. This method, referred to as Continuous Sampling (CS), eliminates the need for disk storage, avoids the bottleneck of data loading, and ensures competitive training times. More importantly, CS provides a continuous stream of fresh images at every training step, maximizing data diversity and effectively addressing the limitations of prior techniques.

\section{Methodology}
\label{sec:method}

\subsection{Measuring the impact of latent variables in MLVGMs}
\label{subsec:mlvgms}

% Definition
Before formalizing our approach for measuring the contribution of single latent variables, we define the concept of Multiple Latent Variable Generative Models (MLVGMs):

%The recent success of Multiple Latent Variable Generative Models (MLVGMs) in diverse applications \citep{stylegan_manipulation,stylegan_translation,stylegan_edit1,stylegan_edit2,serez2024mlvgms} underscores the need for a systematic method to quantify the contribution of each latent variable in the generative process. Developing such a method would enhance our understanding of the hierarchical dynamics of MLVGMs, identifying underutilized or overutilized latent codes and offering valuable insights for optimizing their application in downstream tasks. To formalize our approach, we first define the concept of MLVGMs:

\begin{definition}
    \label{def:mlvgm}
    \emph{(Multiple Latent Variable Generative Models).}
    
    A Multiple Latent Variable Generative Model (MLVGM), denoted $g(\mathbf{z}_1, \mathbf{z}_2, \dots ,\mathbf{z}_{n}; \theta) = \mathbf{x}$, is a deep neural network parameterized by $\theta$. It generates new data $\mathbf{x}$ by modeling $n$ random latent variables $\{ \mathbf{z}_1, \mathbf{z}_2, \dots ,\mathbf{z}_{n} \}$ at different and progressive layers of the network, such that:
    \begin{align*}
        g & :  \mathbb{R}^{m_1} \times \mathbb{R}^{m_2} \times \dots \times \mathbb{R}^{m_{n}}  \rightarrow \mathbb{R}^d \\
        g & :=
        l_{[n]}(\mathbf{z}_{n}, l_{[n-1]}(\mathbf{z}_{n-1}, \dots l_{[1]}(\mathbf{z}_1) \dots));
    \end{align*}
    where $l_{[i]}$ represents the $i^{\text{th}}$ block of the generator, and $\mathbf{z}_i$ is the corresponding latent variable at that layer (parameters $\theta$ are omitted for clarity).
\end{definition}

\paragraph{Intuitions.}

To meaningfully compare the contribution of each latent variable $\mathbf{Z}_i$, we need a metric that operates in the common pixel-space. Probabilistically, we consider images as a random variable $\mathbf{X}$, and therefore

%Each latent variable $\mathbf{z}_i$ contributes differently to the generation of the final output $\mathbf{x}$, depending on its role in the generative process. To compare these contributions meaningfully, we require a metric that operates in a common space. Since the output $\mathbf{x}$ resides in the high-dimensional pixel space and represents a random variable $\mathbf{X}$, we 
select \emph{Mutual Information (MI)} as the metric of choice\footnote{See \Cref{app:theory} for the formal definition of Mutual Information and its probabilistic interpretation.}. 

As an example, let's consider an MLVGM with $n=3$ latent variables, as shown in \Cref{fig:1}. Let $\mathbf{Z}_1, \mathbf{Z}_2, \mathbf{Z}_3$ represent the random latent variables for the three latent spaces, and $\mathbf{X}$ the output in the pixel space. Suppose we perturb the first latent variable, replacing $\mathbf{Z}_1$ with $\hat{\mathbf{Z}}_1$. This generates a modified random variable $\mathbf{X}'$ in the pixel space. We can now relate the average magnitude of the perturbation in the latent space (\eg using $\text{L}_2$ distance), $\mu_1 = \mathbb{E}[\|\hat{\mathbf{Z}}_1 - \mathbf{Z}_1\|_2]$, to the resulting Mutual Information shift in the pixel space, $I(\mathbf{X}, \mathbf{X}') = \gamma$. 

The same process can be repeated for $\mathbf{Z}_2$ and $\mathbf{Z}_3$, introducing $\hat{\mathbf{Z}}_2$ and $\hat{\mathbf{Z}}_3$, and calculating the perturbation magnitudes $\mu_2$ and $\mu_3$ needed to achieve the \emph{same} MI shift $\gamma$ in the pixel space. If the generative process respects the ``global-to-local'' hierarchy typically attributed to MLVGMs (\Cref{fig:1b}), we expect: $\mu_3 > \mu_2 > \mu_1$, as depicted in \Cref{fig:2}. 

Since directly computing MI for high-dimensional variables like $\mathbf{X}$ is analytically intractable, we estimate a lower bound using InfoNCE \citep{InfoNCE}. Additionally, we employ a Monte Carlo procedure to calculate the average perturbations. Details of these computations are provided in the following sections.

\paragraph{Preliminaries.}

InfoNCE loss \citep{InfoNCE} was originally proposed for SSCRL, encouraging similar views (positives) to have close representations, while ensuring that dissimilar views (negatives) remain distant. Formally, it is defined as:

\begin{equation}
    \label{eq:infonce}
        \mathcal{L}_{\text{InfoNCE}} =
        \mathbb{E}_{\mathbf{x}, \mathbf{x}'} \biggl[ - \log 
        \biggl( 
        \frac{
        e^{
        \similarity(f(\mathbf{x}; \phi), f(\mathbf{x}'; \phi)) / \tau
        }
        }
        {
        \sum_{k=1}^K e^{
        \similarity(f(\mathbf{x}; \phi), f(\mathbf{x}^k; \phi)) / \tau
        }
        } 
        \biggr) 
        \biggr];
\end{equation}

where $\mathbf{x}$ and $\mathbf{x}'$ are the anchor and positive images, respectively, $\similarity$ denotes the cosine similarity operator, $f$ is the encoder function parameterized by $\phi$, $\tau$ is a temperature parameter and $K$ is the number of samples (both positive and negative) in a mini-batch.

As demonstrated in \citet{InfoNCE,infobound}, InfoNCE provides a lower bound on the MI between the learned representations:

\begin{equation}
\label{eq:lower_bound_nce}
   \log(2K - 1) - \mathcal{L}_{\text{InfoNCE}}  \le I(f(\mathbf{X}; \phi);  f(\mathbf{X}'; \phi)).
\end{equation}

In typical SSCRL setups (e.g., SimCLR \citep{simclr}), the random variables $\mathbf{X}$ and $\mathbf{X}'$ are generated using deterministic augmentations, such as cropping, flipping, or color adjustment, applied to the same base image. These transformations result in a fixed mutual information value $I(\mathbf{X};  \mathbf{X}')$. Since $f(\cdot; \phi)$ is a deterministic function, the fixed term $I(\mathbf{X};  \mathbf{X}')$ serves as an upper bound to \Cref{eq:lower_bound_nce}, following directly from the data processing inequality (see \Cref{app:theory}):

\begin{equation}
\label{eq:mi_nce_bounds}
    \log(2K - 1) - \mathcal{L}_{\text{InfoNCE}}  \le I(f(\mathbf{X}; \phi);  f(\mathbf{X}'; \phi)) \le I(\mathbf{X};  \mathbf{X}').
\end{equation}

Thus, minimizing the InfoNCE loss in SSCRL can be interpreted as tightening the bounds on mutual information, ensuring that the learned representations effectively capture all relevant information shared between the positive views $\mathbf{X}$ and $\mathbf{X}'$.

\paragraph{The proposed approach.} We build
%\paragraph{Estimating MI in MLVGMs.}
%The proposed algorithm builds 
on the insights of \Cref{eq:mi_nce_bounds}, utilizing InfoNCE as a proxy to measure MI shifts between views. Unlike classical SSCRL methods, which rely on fixed, deterministic transformations, we generate views $\mathbf{X}$ and $\mathbf{X}'$ by perturbing individual latent variables in the latent spaces of a pre-trained MLVGM.

Drawing inspiration from \citet{cop-gen}, which learns latent-space perturbations for positive view generation in LVGMs, our approach adopts an adversarial procedure to optimize InfoNCE loss while progressively reducing the MI between the positive views $\mathbf{X}$ and $\mathbf{X}'$\footnote{A detailed discussion of \citet{cop-gen} is provided in \Cref{subsec:view_gen}.}. 

Formally, let $g$ denote a pre-trained MLVGM with $n$ latent variables and parameters $\theta$. The objective is to identify a perturbation function $T_\mathbf{z}^i(\cdot)$ for each latent space $1 \leq i \leq n$, ensuring that:

\begin{equation}
    I(g(\mathbf{Z}_1, \mathbf{Z}_2, \dots, \mathbf{Z}_i, \dots, \mathbf{Z}_n; \theta); g(\mathbf{Z}_1, \mathbf{Z}_2, \dots, T_{\mathbf{z}}^i(\mathbf{Z}_i), \dots, \mathbf{Z}_n; \theta)) \approx \gamma.
\end{equation}

To achieve this, we model $T_{\mathbf{z}}^i(\mathbf{z}_i)$ as a simple additive perturbation: $T_{\mathbf{z}}^i(\mathbf{z}_i) = \mathbf{z}_i + p(\mathbf{z}_i; \varphi)$, where $p(\cdot)$ is a small multi-layer perceptron (MLP) parameterized by $\varphi$. Since InfoNCE provides the lower bound on MI, we need to compute it by introducing an encoder function $f(\cdot)$ with parameters $\phi$ and define the optimization as a minimax problem (we omit the parameters $\theta$ of the fixed generator $g$):

\begin{equation}
    \label{eq:minmax}
    \max_{\varphi} 
    \min_{ \phi } 
    \mathcal{L}_\text{InfoNCE} \bigl( 
    f(g(\mathbf{z}_1, \mathbf{z}_2, \dots, \mathbf{z}_i, \dots, \mathbf{z}_n); \phi), 
    f(g(\mathbf{z}_1, \mathbf{z}_2, \dots, T_{\mathbf{z}}^i(\mathbf{z}_i; \varphi), \dots, \mathbf{z}_n); \phi) \bigr);
\end{equation}

% Explain how we use it.
\paragraph{Training dynamics.} We initialize parameters $\varphi$ such that $T_{\mathbf{z}}^i(\cdot)$ represents the identity function. In other terms, the applied perturbation is initially zero, and the views $\mathbf{X}$ and $\mathbf{X}'$ are identical. From the perspective of \Cref{eq:mi_nce_bounds}, $I(\mathbf{X};  \mathbf{X}') = H(\mathbf{X})$, corresponding to the trivial setting where the encoder $f$ can achieve $\mathcal{L}_{\text{InfoNCE}} \approx 0$ with ease.
%For the training procedure, we initialize $\varphi$ such that $T_{\mathbf{z}}^i(\cdot)$ represents the identity function. In this state, the perturbation is zero, and the views $\mathbf{X}$ and $\mathbf{X}'$ are identical. As a result, from the perspective of \Cref{eq:mi_nce_bounds}, the mutual information between the views is at its maximum, $I(\mathbf{X};  \mathbf{X}') = H(\mathbf{X})$, which corresponds to a trivial setting where the encoder $f$ can achieve $\mathcal{L}_{\text{InfoNCE}} \approx 0$ with ease. 
As training continues, the perturbation function $T_{\mathbf{z}}^i(\cdot)$ learns to apply progressively larger modifications to the latent variable $\mathbf{Z}_i$, increasing the diversity of the generated views. This, in turn, reduces the mutual information $I(\mathbf{X}; \mathbf{X}')$, thereby \emph{lowering the upper bound} in \Cref{eq:mi_nce_bounds}. As a result, the encoder $f$, tasked with minimizing $\mathcal{L}_{\text{InfoNCE}}$, must maintain the shared information between increasingly distinct views $\mathbf{X}$ and $\mathbf{X}'$ into a common representation, \emph{tightening the lower bound}. 

In summary, $T_{\mathbf{z}}^i$ progressively enhances diversity in the views, reducing $I(\mathbf{X}; \mathbf{X}')$ and causing InfoNCE to increase over time. Conversely, $f$ seeks to learn the most informative representations, tightening the lower bound from the left and seeking equality in \Cref{eq:mi_nce_bounds}. We refer the reader to \Cref{app:training_dynamics} for a detailed graphical illustration of these training dynamics, showing the evolution of InfoNCE loss, average perturbations, and additional insights such as required training time and hyperparameter settings.

\paragraph{Monte Carlo sampling.}
As a result of the above, we obtain $n$ independent perturbation functions $T_{\mathbf{z}_i}(\cdot)$, acting on single latents. For each of these,
%As described above, for an MLVGM with $n$ latent variables, we optimize the minimax objective of \Cref{eq:minmax} independently for each latent variable $i \in \{1, 2, \dots, n\}$. For each $T_{\mathbf{z}_i}(\cdot)$, 
training is stopped when $\mathcal{L}_{\text{InfoNCE}} \approx \overline{\gamma}$, ensuring %that the resulting perturbations produce 
views with a consistent MI shift across all latent variables, and allowing the direct comparison of perturbation magnitudes across different latent spaces.

Afterwards, we perform Monte Carlo (MC) sampling by computing a statistically relevant number of image pairs for each level $i$: $\mathbf{X} = g(\mathbf{Z}_1, \mathbf{Z}_2, \dots, \mathbf{Z}_i, \dots, \mathbf{Z}_n; \theta)$ and $\mathbf{X}' = g(\mathbf{Z}_1, \mathbf{Z}_2, \dots, T_{\mathbf{z}}^i(\mathbf{Z}_i), \dots, \mathbf{Z}_n; \theta)$, where we know that $I(\mathbf{X}; \mathbf{X}') \approx \overline{\gamma}$. This enables to estimate
%After training, we obtain $n$ distinct perturbation functions $T_{\mathbf{z}_i}(\cdot)$, each tailored to one specific latent variable. Using these functions, we generate image pairs $\mathbf{X} = g(\mathbf{Z}_1, \mathbf{Z}_2, \dots, \mathbf{Z}_i, \dots, \mathbf{Z}_n; \theta)$ and $\mathbf{X}' = g(\mathbf{Z}_1, \mathbf{Z}_2, \dots, T_{\mathbf{z}}^i(\mathbf{Z}_i), \dots, \mathbf{Z}_n; \theta)$, such that $I(\mathbf{X}; \mathbf{X}') \approx \gamma$. 
%More specifically, we perform Monte Carlo (MC) sampling by computing a statistically relevant number of views for each level $i$, estimating 
the \textit{average latent perturbation} $\mu_i$ required to achieve a similar MI shift in the image space. As depicted in \Cref{fig:2}, we generally expect that later latent spaces require larger perturbations to achieve the MI shift, matching the empirical observations on the ``global-to-local'' property of MLVGMs.

In \Cref{sec:experiments}, we use this strategy to estimate the impact of latent variables for two distinct MLVGMs: a BigBiGan \cite{bigbigan} pre-trained on ImageNet-$1$K \cite{imagenet} and a StyleGan2 \cite{style-gan2} pre-trained on LSUN Cars \cite{lsun}. The former has $6$ latent variables, while the latter has $16$, which we re-organize into $4$ groups of $4$ for computational practicality. 

%These models are subsequently employed for positive view generation, as described in the following.

\subsection{Positive view Generation Strategies}
\label{subsec:view_gen}

As illustrated in \Cref{fig:3c}, we generate pairs of positive views by applying perturbations to one or more latent spaces, each with an appropriately selected magnitude. 
To better understand the rationale behind our approach, we first look at how previous methods decide view generation strategies.
%The choice is guided by the latent impact estimation procedure described above. However, MI shifts alone (or any other metric) can not establish a systematic method for determining optimal positive views. This limitation arises from the intrinsic complexity of the SSCRL problem, which depends on the specific downstream task and the nature of the data itself, as we elaborate in the following.

\paragraph{Background.}

The problem of Self-Supervised Contrastive Representation Learning (SSCRL) is strictly correlated to designing effective positive views, enabling meaningful representations. In \cite{infomin}, the following principle is introduced:

\begin{proposition}
\label{prop:infomin}
\emph{(Optimal Views for SSCRL, \cite{infomin}).}\\
Given a downstream task $\mathcal{T}$ with labels $\mathbf{Y} \in \mathcal{Y}$, the optimal views $(\mathbf{X^*_1};\mathbf{X^*_2})$ created from data $\mathbf{X}$ are:

\begin{equation}
    \label{eq:infomin}
    (\mathbf{X^*_1};\mathbf{X^*_2}) = \argmin_{\mathbf{X_1};\mathbf{X_2}} I(\mathbf{X_1};\mathbf{X_2}); \text{ subject to } 
    I(\mathbf{X_1};\mathbf{Y}) = I(\mathbf{X_2};\mathbf{Y}) = I(\mathbf{X};\mathbf{Y});
\end{equation}

meaning that the Mutual Information (MI) between optimal views is minimized to contain only the task-relevant information $I(\mathbf{X^*_1};\mathbf{X^*_2}) = I(\mathbf{X};\mathbf{Y})$, while removing all nuisance information, $I(\mathbf{X^*_1};\mathbf{X^*_2}|\mathbf{Y})=0$.
\end{proposition}

The principle states that optimal views should minimize their Mutual Information (MI) while retaining all information relevant to the downstream task, expressed by some label $\mathbf{Y}$. However, in SSCRL, labels are unavailable, and \emph{the downstream task is unknown}. Consequently, designing optimal views becomes infeasible. %As a result, positive view generation methods focus on obtaining broad applicability across various tasks, relying on heuristics or qualitative evaluation rather than a systematic framework.
To solve this inherent shortcoming, most methods design views by implicitly fixing an MI threshold that decides true positives:

\begin{definition}
    \label{prop:mi_consistency}
    \emph{(Mutual Information bound for SSCRL).}\\
    Let $\mathbf{X}_1, \mathbf{X}_2$ be two random variables in the common image space. In absence of a known downstream task $\mathcal{T}$ with labels $\mathbf{Y} \in \mathcal{Y}$; the variables $\mathbf{X}_1, \mathbf{X}_2$ can be considered as positive views if 
    \begin{equation}
      I\bigl(\mathbf{X}_1, \mathbf{X}_2\bigr) \ge \rho;
    \end{equation}
    where $\rho$ is an implicitly defined \textbf{MI threshold}.
\end{definition}

In other terms, the goal of view generation is to define positives that share a \emph{reasonable} amount of information, relevant for as many tasks as possible.
% Examples SIMCLR, gen rep, cop gen
To exemplify this phenomenon, we analyze three prominent positive view generation methods. SimCLR \citep{simclr}, defines a broad set of data augmentations $T_{\mathbf{x}}$ to be applied in the pixel space. The specific transformations and their combinations are selected through ablation studies conducted on the ImageNet-$1$K classification task. In this context, the threshold $\rho$ is implicitly fixed by means of \Cref{eq:infomin}, where a pretext downstream task is considered.%, thereby violating the assumption that the downstream task is unknown. 

In the context of Latent Variable Generative Models (LVGMs), two methods for generating views by perturbing the single latent space stand out. First, \citet{gen-rep} propose applying random perturbations to an anchor latent variable $\mathbf{z}$. Specifically, the perturbation is defined as $T_\mathbf{z}(\mathbf{z}) = \mathbf{z} + \mathbf{w}_{\text{rand}}$, where $\mathbf{w}_{\text{rand}} \sim \mathcal{N}^t(\mu, \sigma, t)$ follows a truncated Gaussian distribution with truncation parameter $t$. Similar to SimCLR, the parameters of the distribution (e.g., the standard deviation $\sigma$) are tuned via ablation studies on ImageNet, fixing $\rho$ with a pretext task. %, again violating the SSCRL assumption. 

In contrast, \citet{cop-gen} introduce an adversarial approach to learn the perturbation $T_{\mathbf{z}}$ for each instance. In this case, the positive view is generated as $T_\mathbf{z}(\mathbf{z}) = \mathbf{z} + \mathbf{w}_{\text{learn}}$, where $\mathbf{w}_{\text{learn}}$ is a learnable perturbation vector. The objective function is formulated similarly to \Cref{eq:minmax}, but applied to LVGMs. In this case, the training stopping criterion (and therefore the threshold $\rho$) is empirically decided by observing the quality of generated views at each step. %the main SSCRL assumption is not violated, as the stopping criterion is decided on empirical inspection of the generated views.

\paragraph{Deciding perturbation magnitudes.}

%In the context of MLVGMs, we 
We maintain the perturbation strategies proposed by \citet{gen-rep} and \citet{cop-gen}, referred to as \textit{random} and \textit{learned}, respectively. 
In fact, the core advantage of our method relies on designing tailored magnitudes based on each latent space's contribution to the generative process, rather than proposing a novel perturbation strategy. 

To do so, we proceed with two distinct phases. First, we use the quantification algorithm to relate the impact of different latents. This reveals the presence of over- or under-used codes, which we can immediately discard. Overused codes imply that even very small perturbations will result in a low $\rho$ threshold, while unused codes do not affect the threshold in any way. Therefore, this first step has the goal to discard perturbations that lead to non-informative views. Finally, in the second phase we consider the remaining codes only and visualize the semantic effect of the applied perturbations to estimate valid magnitudes, implicitly defining $\rho$.

%In fact, our experimental results indicate that the perturbation strategy has a limited impact on the final learned representations. Instead, the \emph{magnitude} of the perturbations plays a far more critical role. As estimated by our proposed algorithm, MLVGMs offer enhanced control over this aspect, enabling us to tailor magnitudes based on each latent space's contribution to the generative process.

%Similarly to prior approaches, however, defining a fully systematic method for selecting the different magnitudes across $n$ latent spaces remains challenging. While MI helps in \emph{quantifying} the impact of each latent variable, it does not provide any information on the \emph{semantic} content of the generated views. To address this limitation, we follow \citet{cop-gen} and incorporate qualitative evaluation into the process, ensuring the SSCRL assumption remains intact.

\begin{figure}[t]
  \centering
  \begin{subfigure}{0.555\linewidth}
   \centering
    \includegraphics[width=0.94\linewidth]{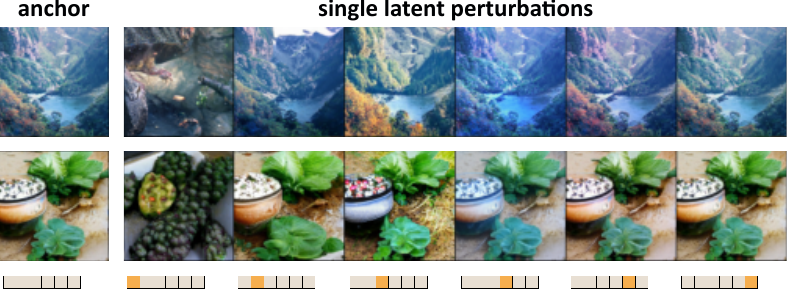}
    \caption{}
    \label{fig:4a}
  \end{subfigure}
  \hfill
  \begin{subfigure}{0.405\linewidth}
    \includegraphics[width=0.94\linewidth]{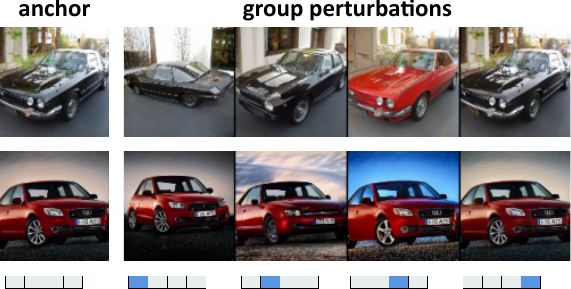}
    \caption{}
    \label{fig:4b}
  \end{subfigure}
  \caption{Examples of views generated by adding the \textbf{same} latent vector $\mathbf{w}$ to different levels. \textbf{(a)} Two anchor images and possible views generated by the perturbations of $6$ BigBiGan's latent levels, represented as the $6$ elements' vector at the bottom. The darker element indicates the applied perturbation $T_\mathbf{z}^i(\mathbf{z}_i) = \mathbf{z}_i + \mathbf{w}$ for each latent level $i$. \textbf{(b)} Generated anchors and views by StyleGan2, which has $16$ hierarchical levels, grouped into $4$ sets and represented as the $4$ elements' vector at the bottom. The darker element indicates the altered group.}
  \label{fig:4}
\end{figure}

As an example, in \Cref{fig:4}
%Specifically, 
we generate multiple examples for each considered MLVGM, by perturbing each latent variable (or group, in the case of StyleGan2) with the same latent vector $\mathbf{w}$. %, as \Cref{fig:4} exemplifies. 
For BigBiGan, the quantitative analysis reveals that the first code is overused, and we can verify that even small perturbations completely change the views semantic content. On the opposite, the last latent is underused, and visually presents no differences w.r.t. the reference image. As a result, we apply perturbations only on the remaining four latents, which act on small shape details or colors.
Similarly, for StyleGan2, we first quantitatively verify that the last group is underused, and discard it. For the remaining ones, we visually observe how the first group controls large-scale transformations, the second group adjusts subject and background composition, while the latter is primarily responsible for color variations.
%For BigBigan, trained on a large range of classes, the first latent variable has a significant semantic impact in the generation process, while subsequent variables influence finer details. 

%In contrast, StyleGan2, which operates on the fine-grained domain of vehicles, exhibits a more balanced sub-division. Specifically, the first group of latent variables controls large-scale transformations, such as rotations and zoom. The second group adjusts subject and background composition, while the final groups are primarily responsible for color variations. In summary, when selecting each perturbation magnitude, we consider both these qualitative observations and the quantitative results measured on MI shifts.

\paragraph{Advantages.}

The inherent disentanglement of coarse and fine-grained features in MLVGMs offers a clear advantage over standard LVGMs, even when employing similar perturbation strategies (\textit{random} or \textit{learned}). To clarify, consider the set of all possible downstream tasks for image data, $\{\mathcal{T}_1, \mathcal{T}_2, \dots, \mathcal{T}_T\}$. For each latent point $\mathbf{z} \sim \mathcal{Z}$, there exists a maximum perturbation magnitude $|\mathbf{w_z}|^{\mathcal{T}_t}$ for each task $\mathcal{T}_t$, such that the resulting views are optimal for the task (\Cref{prop:infomin}).

In SSCRL, however, the downstream task $\mathcal{T}_t$ is unknown. Latent perturbation-based positive view generation methods, therefore, aim to define a function $T_\mathbf{z}(\mathbf{z}) = |\mathbf{w_z}|$ that estimates, for each latent point, a non-trivial perturbation that generates valid (true) positives for as many tasks as possible. In other terms, it exists a direct mapping between the selected latent functions and the choice of the MI threshold $\rho$: while larger $|\mathbf{w}|$ (lower $\rho$)
%While larger $|\mathbf{w}|$ 
can yield more informative (hard) positives, it also increases the likelihood of producing false positives, potentially reducing generalization across diverse tasks. 

The primary advantage of MLVGMs lies in their hierarchical structure, which distributes features across multiple latent spaces. This enables the definition of separate perturbation functions $T_\mathbf{z}^i(\cdot)$ for each latent space $i$, tailored to the impact of that space on the generative process. Crucially, the average perturbation magnitude $|\mathbf{w}|_i$ progressively increases with the latent level index $i$, as illustrated in \Cref{fig:3a}. For example, perturbations in a high-level latent space (responsible for fine details like textures) can be very large without compromising the validity of positive views for most downstream tasks - \ie they have a low effect on $\rho$.

Thus, for an MLVGM with $n$ latent spaces, perturbation magnitudes can be progressively scaled: $|\mathbf{w}|_1 < |\mathbf{w}|_2 < \dots < |\mathbf{w}|_n$. In contrast, LVGMs encode all features within a single latent space, forcing perturbations to be constrained by the most sensitive features. From an MLVGM perspective, this corresponds to a uniform perturbation magnitude, $|\mathbf{w}|_1 = |\mathbf{w}|_2 = \dots = |\mathbf{w}|_n$, which significantly limits flexibility and reduces the impact of generated positive views.

\subsection{Continuous Sampling}

Utilizing generative models to sample both anchor and positive views can degrade final performance \citep{cas}, primarily due to the lower variability of synthetic images compared to real data. To address this limitation, previous methods \citep{supclassifier3,supclassifier7,scaling} have proposed increasing variability by sampling a larger number of images relative to the reference dataset size, ensuring batches are not repeated across epochs. However, the prevailing approach involves sampling this extensive synthetic dataset \emph{offline} (before training), which demands significant storage capacity and additional pre-processing time.

In this study, we avoid these drawbacks by adopting a \emph{Continuous Sampling} strategy that leverages fast generators (such as GANs) to dynamically sample new images during the training of the SSCRL encoder. Specifically, we load the pre-trained generator onto the same GPU device as the encoder and replace the standard data loading step with an on-the-fly generator inference step. This process outputs a new batch of synthetic images directly on the target device, eliminating the need for pre-generated datasets. Since the pre-trained GAN operates exclusively in inference mode, the additional memory overhead is minimal and affordable, allowing us to maintain sufficiently large batch sizes for effective SSCRL training.

With this \emph{Continuous Sampling} approach, the number of training steps per epoch remains consistent with real-data-based training. However, the total number of unique images seen during training is significantly increased, as the effective training set size becomes $n$ \emph{epochs} times larger. Moreover, this strategy eliminates the need for pre-generating and storing extensive datasets and avoids standard data-loading bottlenecks, resulting in training times that are comparable to or faster than traditional methods (see \Cref{sec:experiments}). For a detailed implementation, we provide pseudocode for the continuous sampling procedure in \Cref{app:pseudocode}.

\section{Experiments}
\label{sec:experiments}

In this section, we present the results of our Monte Carlo procedure for quantifying the impact of latent variables on two MLVGMs: BigBiGan and StyleGan2. Subsequently, we utilize these MLVGMs as view generators to train encoders using different SSCRL frameworks, leveraging our proposed Continuous Sampling strategy. 

To evaluate the effectiveness of our approach, we compare it against two existing latent perturbation techniques for LVGMs, specifically those introduced by \citet{gen-rep} and \citet{cop-gen}. As an additional baseline, we include SimCLR, a widely-used view generation method applied to real data, and investigate its combination with transformations applied on top of MLVGM-generated views. Furthermore, in \Cref{app:nvae_experiments}, we extend the applicability of our method to other generative models beyond GANs by training an NVAE \citep{nvae} on the CIFAR-10 dataset \citep{cifar100}.

Finally, we evaluate the overall training efficiency of Continuous Sampling by comparing its runtime performance against standard data loading pipelines, demonstrating its capability to increase data variability without incurring significant computational overhead.

\subsection{Impact of Latent Variables}

\begin{table}[tb]
    \caption{Results of the MC simulation on BigBiGan \textbf{(a)} and StyleGan2 \textbf{(b)}. For each latent level $i$ or group $g$ we show the final InfoNCE loss value and the estimated mean ($\mu$) and standard deviation ($\sigma$) of the corresponding inferred distribution. Average perturbation values confirm that early levels greatly impact the generation process, while later levels may have no impact at all.}
    \label{tab:mc_estimation}
    
    \begin{subtable}{.45\linewidth}
        \centering
        \small
         \caption{}
        \label{tab:mc_bigbigan}
        \begin{tabular}{c |rrr}
        
            \toprule
            
            \multicolumn{1}{c}{latent level} & \multicolumn{1}{c}{loss} & \multicolumn{2}{c}{estimated $q^i$} \\
            \cmidrule(lr){3-4}
            \multicolumn{1}{c}{($i$)} & \multicolumn{1}{c}{(InfoNCE)} &  \multicolumn{1}{c}{($\mu_i$)} & \multicolumn{1}{c}{($\sigma_i$)} \\
            
            \midrule
            $1$ & $1.09$ & $0.67$ & $0.21$ \\
            $2$ & $1.04$ & $3.63$ & $1.18$  \\
            $3$ & $1.05$ & $6.97$ & $1.85$  \\
            $4$ & $1.02$ & $13.00$ & $7.08$  \\
            $5$ & $1.05$ & $21.22$ & $13.68$  \\
            $6$ & $0.14$ & $594.71$ & $616.80$ \\
            \bottomrule
        \end{tabular}
    \end{subtable}
    \hfill
    \begin{subtable}{.45\linewidth}
        \centering
        \small
        \caption{}
        \label{tab:mc_stylegan}
        \begin{tabular}{c |rrrr}
        
            \toprule
            
            \multicolumn{1}{c}{latent group} & \multicolumn{1}{c}{loss} & \multicolumn{2}{c}{estimated $q^g$} \\
            \cmidrule(lr){3-4}
            \multicolumn{1}{c}{($g$)} & \multicolumn{1}{c}{(InfoNCE)} & \multicolumn{1}{c}{($\mu_g$)} & \multicolumn{1}{c}{($\sigma_g$)} \\
           
            \midrule
            $1-4$ & $0.99$ & $15.1$ & $2.7$ \\
            $5-8$ & $1.14$ & $29.0$ & $4.6$  \\
            $9-12$ & $0.94$ & $38.0$ & $5.6$  \\
            $13-16$ & $0.11$ & $134.4$ & $14.2$  \\
            \bottomrule
        \end{tabular}

    \end{subtable} 
\end{table}

Following the procedure detailed in \Cref{subsec:mlvgms}, we train $n$ separate perturbation functions $T_\mathbf{z}^i$ ($n=6$ latent levels for BigBiGan and $n=4$ latent groups for StyleGan2), optimizing the objective in \Cref{eq:minmax}. As visually described in \Cref{app:training_dynamics}, the InfoNCE loss rapidly decreases toward zero during the initial training iterations. As the perturbation functions $T_\mathbf{z}^i$ learn to apply increasing perturbations, the InfoNCE loss rises correspondingly. Training is terminated once a value of $\overline{\gamma} \approx 1$ is achieved, indicating an approximately equal MI shift in the pixel space.

For each latent level or group, we compute the learned perturbation $\mathbf{w_z} = p(\mathbf{z}; \varphi)$ across a statistically significant number of latent points $\mathbf{z}$. This enables us to estimate the mean ($\mu_i$ or $\mu_g$) and standard deviation ($\sigma_i$ or $\sigma_g$) of the inferred perturbation distributions $q^i(|\mathbf{w}|)$ or $q^g(|\mathbf{w}|)$. \Cref{tab:mc_estimation} presents these results, along with the final InfoNCE loss achieved during training. 

From \Cref{tab:mc_bigbigan} (Monte Carlo results for BigBiGan), we observe that the average perturbation (estimated mean $\mu_i$) required to achieve a comparable InfoNCE loss increases progressively across latent levels, from $i=1$ to $i=5$. However, for $i=6$, the InfoNCE loss does not rise substantially even under high average perturbations, suggesting an under-utilization of the latent level in the generative process. Conversely, we measure a very low $\mu_i$ for the first latent level, suggesting a possible over-utilization. These observations may indicate potential inefficiencies in the BigBiGan architecture or training procedure.

A similar trend is observed for StyleGan2 (\Cref{tab:mc_stylegan}), where larger perturbation magnitudes ($\mu_g$) are needed to achieve comparable InfoNCE loss values as latent groups progress from $g=1-4$ to $g=13-16$. Notably, the final group exhibits a degenerate behavior, where even large perturbations fail to influence the MI of the generated views significantly. 

Overall, these results provide clear quantitative evidence that the supposed global-to-local dynamics in MLVGMs hold. Specifically, early latent levels or groups exhibit a stronger influence on the generation process, while later ones primarily affect fine-grained details. To the best of our knowledge, this is the first empirical demonstration of such dynamics across MLVGMs.

\subsection{View Generation}

    % BIGBIGAN
    \begin{table}[tb]
        \centering
        \small
        
         \caption{Comparison of representations learned on the ImageNet-$1$K dataset or BigBiGan generator with two contrastive frameworks (SimCLR and SimSiam). Metrics are Top-$1$ and Top-$5$ accuracy for linear classification on ImageNet-$1$K, average precision for detection on Pascal VOC, and mean Top-$1$ accuracy over $7$ transfer classification datasets. ``random'' row refers to \citet{gen-rep}, and ``learned'' to \citet{cop-gen}. \textbf{Bold} indicates the best result for each group, \underline{\textbf{underline}} the absolute best, and $^*$ indicates the baseline reported from \cite{cop-gen}.} 
        \label{tab:res-bigbigan}
        \setlength\tabcolsep{4.5pt}  % default is 6pt
        \begin{tabular}{l ll | cc ccc | cc ccc c}
            \toprule
            \multirow{3}{*}{Data} & \multirow{3}{*}{$T_{\mathbf{z}}$} & \multicolumn{1}{c}{\multirow{3}{*}{$T_{\mathbf{x}}$}} & \multicolumn{5}{c}{SimCLR} &  \multicolumn{6}{c}{SimSiam} \\
            \cmidrule(lr){4-8} \cmidrule(lr){9-14}

            & & \multicolumn{1}{c}{} & \multicolumn{2}{c}{ImageNet-$1$K} & \multicolumn{3}{c}{Pascal VOC} & \multicolumn{2}{c}{ImageNet-$1$K} & \multicolumn{3}{c}{Pascal VOC} & \multicolumn{1}{c}{Transfer} \\
             \cmidrule(lr){4-5} \cmidrule(lr){6-8} \cmidrule(lr){9-10} \cmidrule(lr){11-13} \cmidrule(lr){14-14}
             & & \multicolumn{1}{c}{} & \multicolumn{1}{c}{Top-$1$} & \multicolumn{1}{c}{Top-$5$} & AP & $\text{AP}_{50}$ & \multicolumn{1}{c}{$\text{AP}_{75}$} &  \multicolumn{1}{c}{Top-$1$} & \multicolumn{1}{c}{Top-$5$} & AP & $\text{AP}_{50}$ & $\text{AP}_{75}$ &  \multicolumn{1}{c}{Top-$1$} \\

            \midrule   
            % real vs synth baselines

            real & - & all & $\mathbf{49.4}^*$ & $\mathbf{75.6}^*$ & $\mathbf{52.9}^*$ & $\mathbf{78.7}^*$ & $\mathbf{58.5}^*$ & \underline{$\mathbf{49.1}$} & \underline{$\mathbf{74.2}$} & \underline{$\mathbf{54.4}$} & \underline{$\mathbf{80.0}$} & \underline{$\mathbf{60.0}$} & 
            $\mathbf{58.2}$ \\
            synth & - & all & $41.6^*$ & $66.6^*$ & $51.0^*$ & $77.2^*$ & $55.8^*$ & $32.2$ & $56.5$ & $51.6$ & $78.2$ & $57.0$ & $47.2$ \\

            \midrule

            % random unitary, random HL, random HL disc
            synth & random & all & $48.7^*$ & $73.1^*$ & $50.2^*$ & $77.0^*$ & $54.4^*$ & $33.4$ & $57.7$ & $51.7$ & $78.4$ & $56.3$ & $47.0$ \\
            synth & ML rand. & no col. & $\mathbf{53.7}$ & $\mathbf{77.2}$ & $\mathbf{53.3}$ & \underline{$\mathbf{79.5}$} & $\mathbf{58.5}$ & $\mathbf{42.5}$ & $\mathbf{67.7}$ & $\mathbf{54.3}$ & $\mathbf{79.9}$ & $\mathbf{59.6}$ & \underline{$\mathbf{59.6}$} \\
            %synth (D) & ML rand. & no col. & - & - & - & - & - & $\mathbf{43.6}$ & $\mathbf{68.6}$ & $54.1$ & $79.8$ & \underline{$\mathbf{60.0}$} & \underline{$\mathbf{60.5}$}\\

            \midrule
            % learned unitary, learned HL, learned HL disc
            synth & learned & all & $53.2^*$ & $77.2^*$ & $53.1^*$ & $78.9^*$ & $58.0^*$ & $33.0$ & $58.2$ & $51.8$ & $78.0$ & $56.7$ & $46.2$\\
            synth & ML learn. & no col. & \underline{$\mathbf{54.4}$} & \underline{$\mathbf{77.9}$} & \underline{$\mathbf{53.4}$} & \underline{$\mathbf{79.5}$} & \underline{$\mathbf{58.9}$} & $\mathbf{39.5}$ & $\mathbf{64.8}$ & $\mathbf{52.5}$ & $\mathbf{78.9}$ & $\mathbf{57.5}$ & $\mathbf{54.9}$ \\
            %synth (D) & ML learn. & no col. & - & - & - & - & - & $\mathbf{40.6}$ & $\mathbf{65.7}$ & $\mathbf{52.9}$ & $\mathbf{79.3}$ & $\mathbf{58.4}$ & $\mathbf{56.4}$\\
            \bottomrule
        \end{tabular}
    \end{table}

     % STYLEGAN
     \begin{table}[tb]
        \centering
        \small

        \caption{Comparison of representations learned on the LSUN Cars dataset or StyleGan 2 generator with two contrastive frameworks (SimSiam and Byol). Metrics are Top-$1$ and Top-$5$ accuracy for linear classification on Stanford Cars and FGVC Aircraft. ``random'' row refers to \citet{gen-rep}, and ``learned'' to \citet{cop-gen}. \textbf{Bold} indicates the best result for each group, \underline{\textbf{underline}} the absolute best.}
        \label{tab:res-stylegan}
        
        \begin{tabular}{l ll | cc cc | cc cc}
            \toprule
            \multirow{3}{*}{Data} & \multirow{3}{*}{$T_{\mathbf{z}}$} & \multicolumn{1}{c}{\multirow{3}{*}{$T_{\mathbf{x}}$}} & \multicolumn{4}{c}{SimSiam} & \multicolumn{4}{c}{Byol}\\ 
            
            \cmidrule(lr){4-7} \cmidrule(lr){7-11} 
            
             \multicolumn{3}{c}{} & \multicolumn{2}{c}{Stanford Cars} & \multicolumn{2}{c}{FGVC Aircraft} & \multicolumn{2}{c}{Stanford Cars} & \multicolumn{2}{c}{FGVC Aircraft}\\
             
             \cmidrule(lr){4-5} \cmidrule(lr){6-7} \cmidrule(lr){8-9} \cmidrule(lr){10-11}
             
             \multicolumn{3}{c}{} & Top-$1$ & \multicolumn{1}{c}{Top-$5$} & \multicolumn{1}{c}{Top-$1$} & Top-$5$ & Top-$1$ & \multicolumn{1}{c}{Top-$5$} & \multicolumn{1}{c}{Top-$1$} & Top-$5$ \\

             % WITHOUT PM
            \midrule
            real & - & all & $\mathbf{33.4}$ & $\mathbf{64.3}$ & $20.7$ & $48.8$ & $\mathbf{48.9}$ & $\mathbf{79.3}$ & \underline{$\mathbf{35.0}$} & \underline{$\mathbf{65.6}$}\\
            synth & - & all & $27.0$ & $54.6$ & $\mathbf{21.3}$ & $\mathbf{50.5}$ & $40.5$ & $69.6$ & $31.2$ & $61.9$\\
            
            \midrule
            synth & random & all & $29.2$ & $58.1$ & $22.5$ & $51.7$ & $44.6$ & $73.3$ & $30.5$ & $60.4$\\
            synth & ML rand. & no col. & \underline{$\mathbf{47.0}$} & \underline{$\mathbf{76.1}$} & $\mathbf{22.9}$ & \underline{$\mathbf{53.5}$} & \underline{$\mathbf{58.7}$} & \underline{$\mathbf{84.8}$} & $\mathbf{32.5}$ & $\mathbf{61.8}$\\
            
            \midrule
            synth & learned & all & $28.6$ & $56.7$ & $22.0$ & $51.9$ & $45.6$ & $73.6$ & $\mathbf{31.7}$ & $\mathbf{62.1}$ \\
            synth & ML learn. & no col. & $\mathbf{35.2}$ & $\mathbf{64.8}$ & \underline{$\mathbf{23.0}$} & $\mathbf{53.0}$ & $\mathbf{47.8}$ & $\mathbf{77.1}$ & $30.7$ & $61.1$ \\
            \bottomrule

            % WITH PM
            % \midrule
            % real & - & all & $\mathbf{33.4 \pm 0.8}$ & $\mathbf{64.3 \pm 0.4}$ & $20.7 \pm 0.4$ & $48.8 \pm 1.1$ & $\mathbf{48.9 \pm 0.1}$ & $\mathbf{79.3 \pm 0.1}$ & \underline{$\mathbf{35.0 \pm 0.2}$} & \underline{$\mathbf{65.6 \pm 0.6}$}\\
            % synth & - & all & $27.0 \pm 0.2$ & $54.6 \pm 0.2$ & $\mathbf{21.3 \pm 0.7}$ & $\mathbf{50.5 \pm 0.8}$ & $40.5 \pm 0.1$ & $69.6 \pm 0.1$ & $31.2 \pm 0.4$ & $61.9 \pm 0.4$\\
            
            % \midrule
            % synth & random & all & $29.2 \pm 0.4$ & $58.1 \pm 0.2$ & $22.5 \pm 0.6$ & $51.7 \pm 0.7$ & $44.6 \pm 0.1$ & $73.3 \pm 0.1$ & $30.5 \pm 0.4$ & $60.4 \pm 0.2$\\
            % synth & ML rand. & no col. & \underline{$\mathbf{47.0 \pm 0.3}$} & \underline{$\mathbf{76.1 \pm 0.3}$} & $\mathbf{22.9 \pm 0.8}$ & \underline{$\mathbf{53.5 \pm 0.8}$} & \underline{$\mathbf{58.7 \pm 0.1}$} & \underline{$\mathbf{84.8 \pm 0.1}$} & $\mathbf{32.5 \pm 0.4}$ & $\mathbf{61.8 \pm 0.3}$\\
            
            % \midrule
            % synth & learned & all & $28.6 \pm 0.5$ & $56.7 \pm 0.2$ & $22.0 \pm 1.1$ & $51.9 \pm 0.5$ & $45.6 \pm 0.3$ & $73.6 \pm 0.2$ & $\mathbf{31.7 \pm 0.3}$ & $\mathbf{62.1 \pm 0.1}$ \\
            % synth & ML learn. & no col. & $\mathbf{35.2 \pm 0.4}$ & $\mathbf{64.8 \pm 0.5}$ & \underline{$\mathbf{23.0 \pm 1.1}$} & $\mathbf{53.0 \pm 0.9}$ & $\mathbf{47.8 \pm 0.1}$ & $\mathbf{77.1 \pm 0.1}$ & $30.7 \pm 0.5$ & $61.1 \pm 0.2$ \\
            % \bottomrule
            
        \end{tabular}
        
    \end{table}
    
    We test MLVGMs generated views by training multiple ResNet-$50$ encoders, using SimSiam \citep{simsiam}, SimCLR \citep{simclr} (on BigBiGan, following previous work \citet{cop-gen}) and Byol \citep{byol} (on StyleGan2). We sample latent anchors from a truncated normal distribution: $\mathcal{N}^t(0.0, 1.0, 2.0)$ for BigBiGan and $\mathcal{N}^t(0.0, 1.0, 0.9)$ for StyleGan2. Positives are computed using the \textit{random} or \textit{learned} strategies, applied separately on each latent level. Given the MC results reported in \Cref{tab:mc_estimation}, we first select the latents to discard. These are the first latent of BigBiGan (overused) and the last latent/group in both BigBiGan and StyleGan2 (underused). Then, we systematically apply different perturbations to the remaining latents, observing that latents $2-5$ in BigBiGan obtain relevant semantic changes with a similar perturbation magnitude. For StyleGan2, we apply only tiny perturbations on the first two groups (modify shape and global transformations), and a larger perturbation on the remaining group (alters colors).
    %Given the views visualization of \Cref{fig:3} and the MC results reported in \Cref{tab:mc_estimation}, we fix the first latent level on BigBiGan and apply only tiny perturbations on the first two groups of StyleGan2, in order to not affect sensible semantic aspects. Conversely, we enhance the perturbations of the remaining levels, which alter more local details/colors. 
    The specific magnitudes, as well as other hyperparameters, are reported in \Cref{app:training_details}.

    The representation capabilities of the obtained encoders are compared against several methods: training on synthetic data without latent perturbations $T_{\mathbf{z}}$, the \textit{random} and \textit{learned} baselines using single latent spaces, and the upper bound of using real data ($1.28$M images for ImageNet-$1$K \citep{imagenet} and $893$K images for LSUN Cars \citep{lsun}). In all these scenarios, SimCLR pixel-space augmentations $T_\mathbf{x}$ are used, consisting of random cropping, horizontal flipping, grayscale, and color jittering. Since our ML views generate realistic color changes (see \Cref{app:qualitative}), we only partially apply $T_\mathbf{x}$ transformations on top of our positives, removing grayscale and color jittering. To better investigate this aspect, in \Cref{app:ablations} we further test various combinations of $T_\mathbf{x}$ coupled with our method. 
    
    BigBiGan views are evaluated on ImageNet-$1$K linear classification and, for Simsiam, on seven transfer datasets: Birdsnap \citep{birdsnap}, Caltech$101$ \citep{caltech101}, Cifar$100$ \citep{cifar100}, DTD \citep{DTD}, Flowers$102$ \citep{flowers102}, Food$101$ \citep{food101}, and Pets \citep{pets}. We also compute Average Precision on Pascal VOC \citep{pascal-voc} object detection using \verb|detectron 2| \citep{detectron2} to train a Faster-RCNN with the R$50$-C$4$ backbone. The results are reported in \Cref{tab:res-bigbigan}, including the mean accuracy for the transfer tasks (complete runs in \Cref{app:transfer}). For StyleGan2, we compute linear classification accuracy on Stanford Cars \citep{stanford-cars} and FGCV Aircraft $2013$b \citep{aircraft}, reporting results in \Cref{tab:res-stylegan}.

    In all experiments, MLVGMs views outperform the corresponding baseline, proving their superior quality. Comparing \textit{random} and \textit{learned} methods, we observe that the multiple latent (ML) \textit{random} experiments often close the gap with the \textit{learned} counterparts. This suggests that distinct-level perturbations are more important than the selected alteration technique. In comparison with real data, ML views generally yield better or similar results, except in the case of SimSiam encoders evaluated on ImageNet-$1$K. However, this gap narrows or disappears in other downstream tasks and datasets, evidencing good generalization capabilities of the learned representations, which is the main goal of SSCRL. For StyleGan2, the great performance boost given by ML \textit{random} views on Stanford Cars is noteworthy. When generalizing to FGCV Aircraft, all runs achieve similar performance, with marginal improvements of the ML runs when using SimSiam, and good real data results on Byol. This may be due to the high domain shift between the two datasets (Car vs Aircraft), leading to a challenging generalization for all representations.

    \subsection{Continuous Sampling} 

     \begin{figure}[tb]
      \centering
      \includegraphics[width=0.98\linewidth]{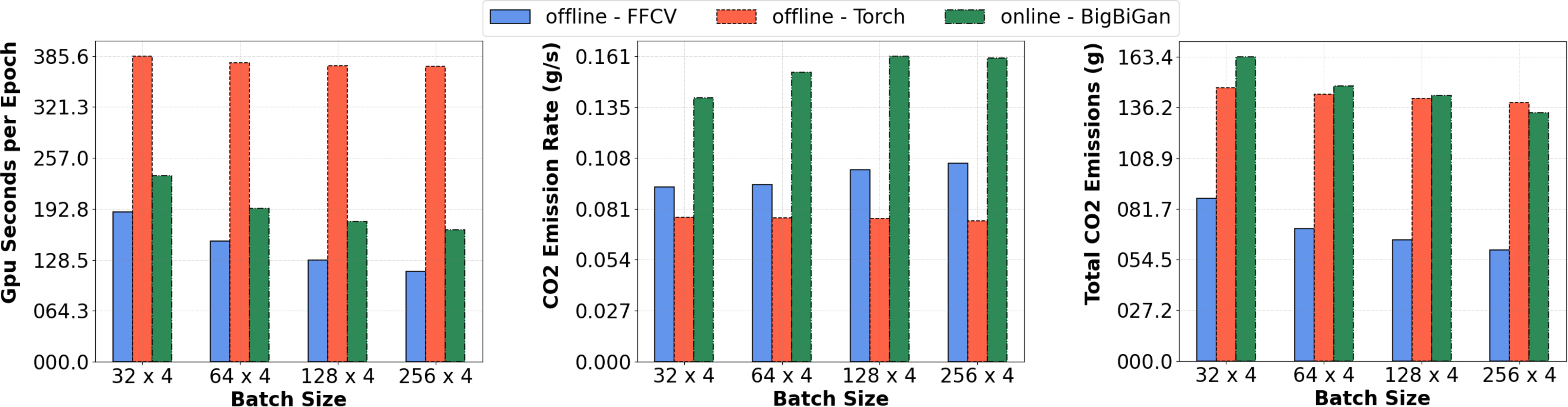}
      \caption{Total time (GPU seconds), $\text{CO}_2$ emissions rate (grams per second) and total $\text{CO}_2$ emissions (grams) for the three tested data loading procedures and different batch sizes.}
      \label{fig:5}
    \end{figure}
    
    All our encoders are trained using Continuous Sampling, except for SimCLR, which follows the previous setup. Additionally, to compare overall training speed to standard data loading, we trained a ResNet-$18$ model with SimCLR for $20$ epochs on ImageNet-$100$, on $4$ NVIDIA A100-SXM4-40GB GPUs and different batch sizes ($32 \times 4$, $64 \times 4$, $128 \times 4$, $256 \times 4$). The experiment compares the standard \verb|PyTorch| \cite{pytorch} loader, the efficient \verb|FFCV| \cite{ffcv} loader (both with $8$ workers), and the BigBiGan generator. \Cref{fig:5} displays our findings, reporting the mean GPU seconds per epoch, the $\text{CO}_2$ emissions rate, and the total $\text{CO}_2$ emissions estimated using \verb|CodeCarbon| \cite{codecarbon}. Continuous Sampling proves significantly faster than \verb|torch| loader and only marginally slower than \verb|FFCV|. In terms of $\text{CO}_2$ emissions rate, the use of BigBiGan led to higher energy consumption, due to intensive GPU usage. Nevertheless, in terms of total $\text{CO}_2$ the values remain comparable with \verb|torch|. In conclusion, the increasing efficiency and precision of modern image generation models, especially fast-sampling GANs, make Continuous Sampling an interesting alternative to conventional data-loading techniques, allowing great image variability while maintaining comparable training times. 
    
\section{Discussion and Conclusions}
\label{sec:conclusions}

In this paper, we explored the influence of multiple latent spaces in MLVGMs' image generation, quantifying their impact as MI shifts in the common pixel space. This approach advances beyond previous empirical observations, providing deeper insights into the generative process, revealing under- or over-utilized latent variables, and guiding the use of MLVGMs in downstream applications. Additionally, we expanded the use of MLVGMs to a new downstream task, which is positive view generation for SSCRL, demonstrating superior results \wrt previous methods using single-variable models and competing with real data training. We also introduced Continuous Sampling, which allows using generators as a data source, creating large training sets without requiring significant storage capacity and achieving comparable or faster training times than standard data loading.

\paragraph{Limitations and impact.} 

Our work showcases MLVGMs as a distinct category of models, offering new tools to assess the impact of latent variables. Specifically, the proposed Monte Carlo quantification method supports previous empirical observations on the ``global-to-local'' nature of MLVGMs, but allows a more in-depth and quantitative analysis. Nonetheless, our algorithm remains limited by the nature of Mutual Information itself, which is invariant under invertible transformations. As a consequence, some small perturbations (e.g. a constant scaling to each pixel), will be approximated as a 0 shift by our method. However, since the differences in average perturbation between each latent space are usually significant, it is reasonable to assume that such small approximations do not invalidate our findings. More specifically, we reveal that modern gan-based MLVGMs, such as BigGan and StyleGan employ over or under-utilized variables in the generative process, setting up a base for possible architectural improvements. In terms of view generation, our method has proven its superiority, surpassing previous perturbation strategies applied to single-variable models. However, it does not address the inherent challenge of SSCRL: views are defined upon ``reasonable'' thresholds, since ``optimal'' positives depend on the specific downstream task. Regarding generative models as a data source, they offer potential solutions to issues associated with real datasets, such as privacy concerns and usage rights \cite{privacy-medical01,privacy-medical02}. However, generative models can inherit biases from the original data \cite{data-bias}, and latent perturbations may amplify them, propagating to downstream models. Therefore, techniques to mitigate these biases could be considered \cite{gen-debias01,gen-debias02}. As for Continuous Sampling, while it reduces disk usage and data diversity, scaling to high image resolutions or large models may be very GPU-intensive and result in high CO2 emissions, limiting usability.

\newpage
\appendix

\section{Information Theory}
\label{app:theory}

\paragraph{Mutual Information.}

Mutual Information (MI) measures the amount of information that one random variable contains about another. More specifically, it measures the reduction in the uncertainty of one random variable due to the knowledge of the other.

\begin{definition} Mutual Information (MI).
\label{def:mutual_information}

Let $\mathbf{X}$ and $\mathbf{Y}$ be two random variables with joint probability $p(\mathbf{x}, \mathbf{y})$ and marginals $p(\mathbf{x})$ and $p(\mathbf{y})$. The Mutual Information (or MI in short) $I(\mathbf{X};\mathbf{Y})$ is the Relative Entropy (or Kullback-Leibler divergence) between the joint distribution and the product of marginals distribution:

\begin{align*}
I(\mathbf{X};\mathbf{Y}) & = D_{\text{KL}}(p(\mathbf{x}, \mathbf{y}) || p(\mathbf{x}) p(\mathbf{y})) \\
& = \int_{\mathbf{x} \in \mathcal{X}} \int_{\mathbf{y} \in \mathcal{Y}} p(\mathbf{x}, \mathbf{y}) \log \frac{p(\mathbf{x}, \mathbf{y})}{p(\mathbf{x}) p(\mathbf{y})}.
\end{align*}

\end{definition}

Note that $I(\mathbf{X};\mathbf{Y}) \ge 0$, with equality iff $\mathbf{X}$ and $\mathbf{Y}$ are conditionally independent.

\paragraph{Data Processing Inequality.}

Formally, the data processing inequality can be formulated as: 

\begin{theorem} Data Processing Inequality.
\label{th:dpi}

If three random variables $\mathbf{X}, \mathbf{Y}, \mathbf{Z}$ form a Markov Chain ($\mathbf{X} \rightarrow \mathbf{Y} \rightarrow \mathbf{Z}$), then:
\[
I(\mathbf{X}; \mathbf{Y}) \ge I(\mathbf{X}; \mathbf{Z}), 
\]
with equality iff $I(\mathbf{X}; \mathbf{Y}|\mathbf{Z}) = 0$. In other terms, no processing of $\mathbf{Y}$, deterministic or random, can increase the information that $\mathbf{Y}$ contains about $\mathbf{X}$.
\end{theorem}

\section{Training Dynamics}
\label{app:training_dynamics}

\begin{figure}[t]
    \begin{center}
        \begin{subfigure}{0.50\linewidth}
            \includegraphics[width=1.00\linewidth]{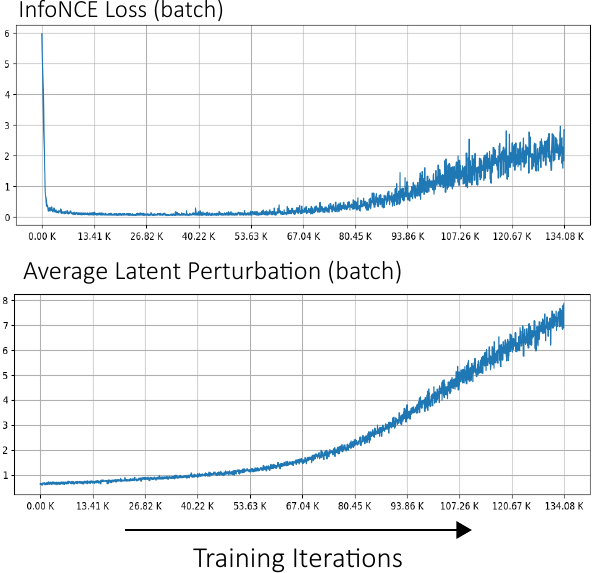}
            \caption{}
            \label{fig:B1a}
        \end{subfigure}
        \hfill
        \begin{subfigure}{0.48\linewidth}
            \includegraphics[width=1.00\linewidth]{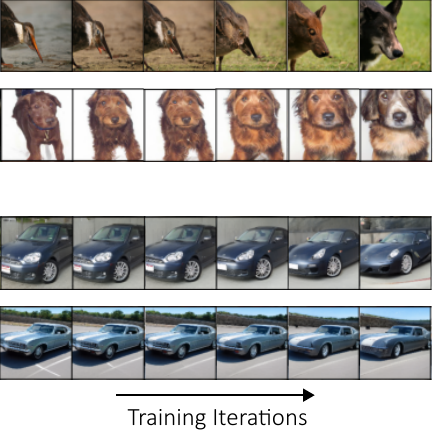}
            \caption{}
            \label{fig:B1b}
        \end{subfigure}
    \end{center}

  \caption{\textbf{(a)} Example of training dynamics comparing the InfoNCE Loss (top) and the average latent perturbation (bottom) for a single batch. $X$ axis denotes training iterations. \textbf{(b)}. Qualitative examples of the generated views' evolution during training (anchor view on the left).}
  
  \label{fig:B1}
\end{figure}

As explained in \Cref{subsec:mlvgms}, we train each perturbation function $T_{\mathbf{z}}^i$ separately, and initialize as the identity function. This allows InfoNCE loss in \Cref{eq:minmax} to start from a very low value. During training, $T_{\mathbf{z}}^i$ progressively enhances diversity in the views, reducing $I(\mathbf{X}; \mathbf{X}')$, and causing InfoNCE to increase over time. An example of these training dynamics is illustrated in \Cref{fig:B1a}. \Cref{fig:B1b} shows some qualitative examples of how views in the pixel space evolve during training. Specifically, the image on the left represents the anchor, while other images show the corresponding perturbed view as the magnitude increases.

For identity initialization, we follow a Gaussian distribution $\mathcal{N}(0.0, 0.01)$ for the weights and a Uniform distribution $\mathcal{U}(-0.001, 0.001)$ for the biases, implying  $T_\mathbf{z}^i(\mathbf{z}) \approx \mathbf{z}$. We use a batch size of $64$, Adam \citep{adam} optimizer with $\beta_1 = 0.5, \beta_2 = 0.999$ and a temperature $\tau = 0.1$ in the InfoNCE loss \citep{InfoNCE}. 

Regarding training time, each $T_{\mathbf{z}}^i$, a two-layer MLP, requires approximately two hours of training on one A-100 GPU. At inference time, we measure that Monte Carlo sampling of 100K latents takes no more than $100$ seconds per latent level, using an RTX $2080$ TI GPU.
        
\section{Continuous Sampling}
\label{app:pseudocode}

We detail the batch generation procedure for anchors and positive views in \Cref{alg:continuous_sampling}. As explained in the main paper, the algorithm requires loading the pre-trained generative model (MLVGM in our case) on the same GPU device as the encoder model to be trained. In this way, the generated batch will be already on the correct device, eliminating the need for data loading. 

\begin{algorithm}[t]
\caption{Continuous Sampling of Batch (training time)}
\label{alg:continuous_sampling}

\textbf{Input}: MLVGM model $g$ with $n$ latent variables; anchors sampling function $N^t(\mu, \sigma, t)$; positive perturbation functions (\textit{random} or \textit{learned}) $T^i_{\mathbf{z}}$. \\
\textbf{Parameter}: batch size $B$. \\
\textbf{Output}: $\mathcal{A}, \mathcal{P}$ Generated batch of anchors and positives.\\

\begin{algorithmic}[1]
\State $\mathcal{A} \gets \emptyset$ \Comment{initialize empty anchors set}
\State $\mathcal{P} \gets \emptyset$ \Comment{initialize empty positives set}

\For{$b \in \{0, \dots, B - 1\}$}
    \State $A_b \gets \emptyset$ \Comment{initialize empty set of anchor variables}
    \State $P_b \gets \emptyset$ \Comment{initialize empty set of positive variables}

    \For{$i \in \{0, \dots, n - 1\}$}
        \State $a_i \sim N^t(\mu, \sigma, t)$ \Comment{sample anchor code}
        \State $p_i \gets T_\mathbf{z}^i(a_i)$ \Comment{generate positive code}
        \State $A_b = A_b \cup \{a_i\}$
        \State $P_b = P_b \cup \{p_i\}$ 
    \EndFor

    \State $\mathbf{a}_b \gets g(A_b)$ \Comment{generate anchor image}
    \State $\mathbf{p}_b \gets g(P_b)$ \Comment{generate positive image}
    \State $\mathcal{A} = \mathcal{A} \cup \{\mathbf{a}_b\}$
    \State $\mathcal{P} = \mathcal{P} \cup \{\mathbf{p}_b\}$ 
\EndFor

\State \Return $\mathcal{A}, \mathcal{P}$ \Comment{SSCRL Encoder's inputs}
\end{algorithmic}
\end{algorithm}

\section{View Generation Training Details}
\label{app:training_details}

The code for this work has been developed using two popular \verb|python|'s deep learning libraries: \verb|pytorch| \citep{pytorch} and \verb|pytorch lightning| \citep{pytorch-lightning}. The BigBiGan generator's code and weights have been introduced in \citep{unsupervisedgan}, and can be obtained at \citep{pytorch-pretrained-gans}. For StyleGan2, the official \verb|github| repositories are available, specifically for code \citep{stylegan2-ada-pytorch} and weights \citep{stylegan2}. In the following, we report the training-specific details used in the implementation.

\paragraph{Training and evaluating encoders.}

    We train each SSCRL encoder (regardless of the framework) for $100$ epochs using SGD optimizer with momentum $0.9$ and weight decay $1 \times 10^{-4}$. The learning rate is set as $0.05 \times \text{BatchSize} / 256$, with a cosine decay scheduler and an additional linear warmup for the first $10$ epochs if $\text{BatchSize} \ge 1024$. Specifically, the BigBiGan encoders are trained with a batch size of $1024$, while for StyleGan2 we use a batch size of $512$, due to the generator's larger number of parameters. Matching the output resolution of BigBiGan and following prior studies \citep{gen-rep,cop-gen}, all views have $128 \times 128$ image resolution.
    
    For sampling positives, the \textit{random} baselines are trained with $\mathcal{N}^t(0.0, 0.2, 2.0)$ and $\mathcal{N}^t(0.0, 0.25, 0.9)$ for each generator, respectively BigBiGan and StyleGan2. Following the suggestions of the original work \citep{cop-gen}, we monitor the generated views and select the best checkpoints for the \textit{learned} baselines. For our ML views, BigBiGan samples positives by fixing the first latent and with parameters $\mathcal{N}^t(0.0, 1.0, 2.0)$ for the remaining codes (including for simplicity the unused latent). On the other hand, we perturb the StyleGan2 selected groups with parameters: $\mathcal{N}^t(0.0, 0.2, 1.0)$; $\mathcal{N}^t(0.0, 0.1, 1.0)$; $\mathcal{N}^t(0.0, 0.8, 1.0)$; $\mathcal{N}^t(0.0, 0.8, 1.0)$, respectively. Similar to baselines, the \textit{learned} perturbations are decided by monitoring checkpoints. In general, they follow the same perturbation magnitudes as the \textit{random} counterpart.
    
    All linear classifiers used for BigBiGan evaluation are trained for $60$ epochs with a batch size of $256$, SGD optimizer, and a learning rate of $30.0$ with cosine decay. StyleGan classifiers follow the same setup, but they are trained for $100$ epochs. For Pascal VOC detection, the R$50$-C$4$ backbone is fine-tuned for $24000$ iterations on \verb|trainval07+12| split and evaluated on \verb|test07|.

\paragraph{Data and preprocessing.}

    All our experiments use \verb|FFCV| \citep{ffcv} library for efficient data storage and fast loading. ImageNet-$1$K's images are stored at $256 \times 256$ resolution and resized to $128 \times 128$ during loading, to match the output resolution of BigBiGan. Regarding LSUN Cars/StyleGan2, instructions to download the $893$K training images can be found at \citep{stylegan2-ada-pytorch}. These are $512 \times 384$ resolution images, which are stored at $512 \times 512$ with black padding, to match StyleGan2 outputs. During loading, images are first center cropped at $384 \times 384$, removing padding, and then resized at $128 \times 128$. The same preprocessing is applied to the generated images.
    
    The data augmentation and preprocessing pipelines rely on the \verb|kornia| library \citep{kornia}. During transfer classification learning, we apply random resize crop and random horizontal flip during training, and center crop during validation/testing. In all experiments, images are normalized with ImageNet mean and standard deviation values, and the final size (after cropping) is $112 \times 112$.

\paragraph{Hardware resources and reproducibility.}

    Most experiments have been run using $4$ NVIDIA A$100$-SXM$4$-$40$GB GPUs, with an exception for the StyleGan2-trained encoders, which required $8$ GPUs of the same type, due to the larger number of parameters of StyleGan2 \wrt BigBiGan. Other minor experiments, like the training of perturbation functions, required only $1$ GPU. To ensure reproducibility, random seeds have always been fixed. For Continuous Sampling, seeds are changed at every iteration and device-specific. This procedure avoids reproducing the same batches during training but allows for consistency throughout different runs.
    
\paragraph{Perturbations in $\mathcal{Z}$ and $\mathcal{W}$ space.}

    In BigBiGan experiments, perturbations in the latent space are applied by summing the noise vector to the selected latents, as described in \Cref{sec:method} of the main paper. For StyleGan2, the final latent space is known as $\mathcal{W}$, and a mapping network is used to perform $f(\mathbf{z}) = \mathbf{w}$. Here, $f$ is the mapping network, $\mathbf{w}$ the random latent vector in the $\mathcal{W}$ space, and $\mathbf{z}$ the initial random vector sampled from a truncated Gaussian distribution ($\mathcal{Z}$ space). The positive views in this case are obtained as $f(T_\mathbf{z}(\mathbf{z})) = \mathbf{w}$, where  $T_\mathbf{z}$ is a \textit{random} or \textit{learned} perturbation that affects only the selected variables.
        
\section{Ablation Studies}
\label{app:ablations}

 \begin{figure}[tb]
    \centering
    \includegraphics[width=0.8\linewidth]{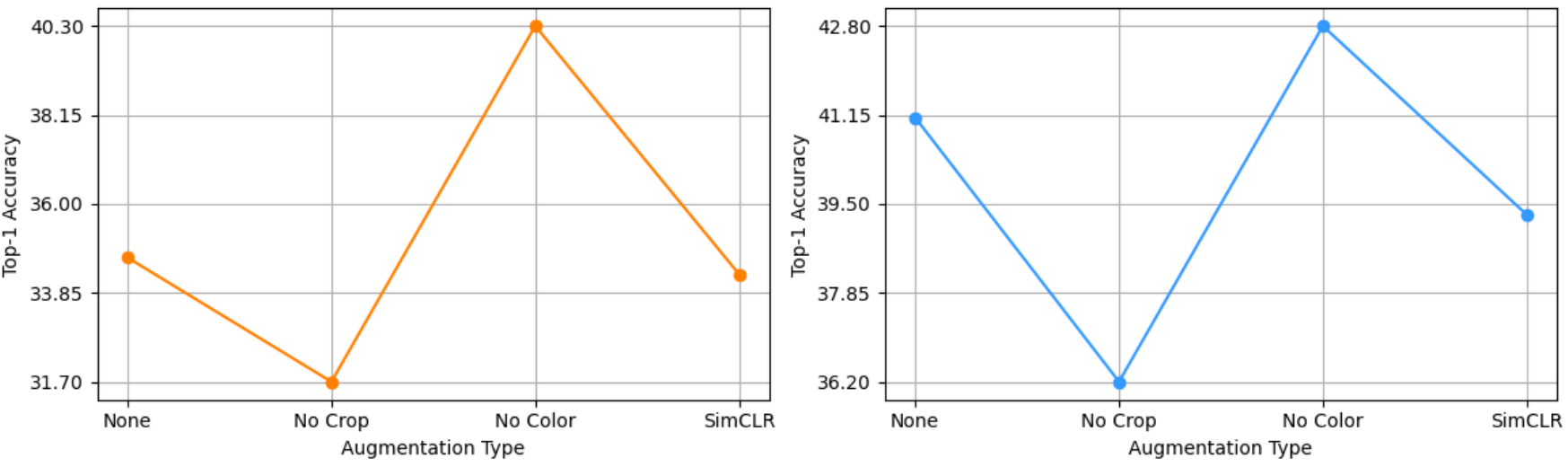}
    \caption{Ablation study on the $T_\mathbf{x}$ pixel-space augmentations combined with BigBiGan's generated views. Left (orange) represents \emph{random} generated views while right (blue) represents \emph{learned} generated views.}
    \label{fig:E_1a}
\end{figure}

\begin{figure}[tb]
    \centering
    \includegraphics[width=0.8\linewidth]{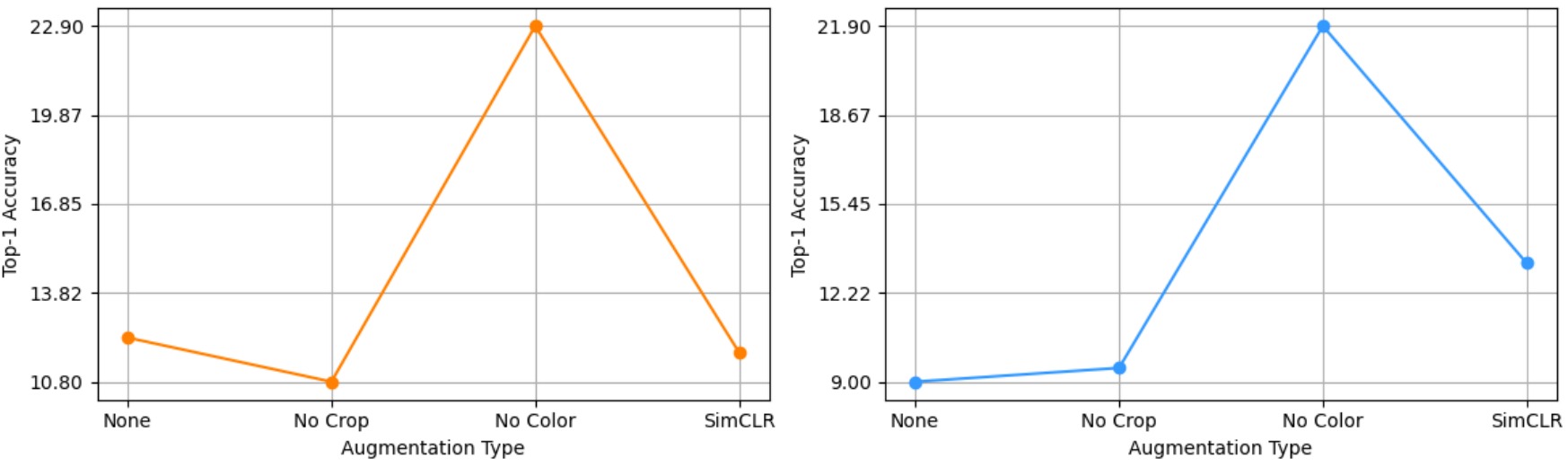}
    \caption{Ablation study on the $T_\mathbf{x}$ pixel-space augmentations combined with StyleGan2's generated views. Left (orange) represents \emph{random} generated views while right (blue) represents \emph{learned} generated views.}
    \label{fig:E_1b}
\end{figure}

Previous studies \citep{gen-rep,cop-gen} show that the \textit{random} and \textit{learned} baselines still benefit from the application of $T_\mathbf{x}$ SimCLR augmentations on top of the generated views. In detail, they perform some ablation studies training encoders on Imagenet-$100$ and a subset of LSUN cars, where different combinations of pixel-space augmentations are tested: cropping and horizontal flipping, grayscale and color jittering, none of the previous or all of them. When utilizing ML perturbations, we qualitatively observe that the generated views can assume various realistic colors (as seen for example in \Cref{fig:4} of the main paper). For this reason, in the experimental analysis, we only use cropping and flipping as pixel space augmentations.

To validate our hypothesis, in \Cref{fig:E_1a,fig:E_1b} we report the result of the same experiments of previous work, but performed on our ML views. In all cases, removing color jittering and grayscale operations increases the final linear evaluation Top-1 accuracy. These observations empirically confirm that ML views do not need additional color transformations.

\section{NVAE Experiments}
\label{app:nvae_experiments}

In the main paper, experiments are conducted on two state-of-the-art GAN methods. To showcase the general properties of the proposed approach, we conducted an additional ablation study where we trained a small NVAE \citep{nvae} with $24$ latent variables, organized in $3$ groups, on
%$3$ latent scales on 
Cifar$10$ \citep{cifar100} ($32\times32$ resolution). The model achieves a FID of $23.63$, and an IS score of $6.36 \pm 0.16$. 

At first, we measure the impact of each latent group replicating the experiments of \Cref{tab:mc_estimation}. The results, included in \Cref{tab:mc_nvae}, confirm that the ``global-to-local'' trend is valid also in this case. Moreover, we observe that the first group is highly influencing the final outcome with respect to the remaining groups.

\begin{table}
    \centering
    \small
     \caption{Results of the MC simulation on the NVAE model pre-trained on Cifar-10. For each latent group $g$ we show the final InfoNCE loss value and the estimated mean ($\mu$) and standard deviation ($\sigma$) of the corresponding inferred distribution.}
    \label{tab:mc_nvae}
    \begin{tabular}{c |rrr}
    
        \toprule
        
        \multicolumn{1}{c}{latent group} & \multicolumn{1}{c}{loss} & \multicolumn{2}{c}{estimated $q^i$} \\
        \cmidrule(lr){3-4}
        \multicolumn{1}{c}{($g$)} & \multicolumn{1}{c}{(InfoNCE)} &  \multicolumn{1}{c}{($\mu$)} & \multicolumn{1}{c}{($\sigma$)} \\
        
        \midrule
        $1$ & $0.60$ & $0.04$ & $1e-3$ \\
        $2$ & $0.60$ & $0.96$ & $0.41$  \\
        $3$ & $0.60$ & $1.18$ & $1.25$  \\
        \bottomrule
    \end{tabular}
\end{table}

Afterwards, we test the model's 
%To test its 
abilities as a data source, %we replicated 
replicating the experiments of \Cref{tab:res-bigbigan} for the \textit{random} case with SimSiam framework and Continuous Sampling. \Cref{tab:nvae_ablation} shows the final Top-$1$ and Top-$5$ accuracies obtained on the Cifar$10$ evaluation set.
In this case, the absolute best result is achieved by the same encoder trained on real data ($79.15\%$ accuracy), with a large gap ($54.95 \%$) when using NVAE as a data source without further latent perturbations. Adding the same perturbations to all groups provides an immediate accuracy boost ($+5\%$), with various standard deviations. After observing the results in \Cref{tab:mc_nvae}, we decided to fix the first group (overused), and find that applying larger perturbations to groups 2-3 only allows for a larger std and additional boost, with a final accuracy of $60.75 \%$. These experiments prove the feasibility of our methods also on Autoencoder-based MLVGMs, despite the small model employed.
%, where the parameters for sampling anchors have been learned during training of the NVAE. \textbf{(a)}: three different perturbation magnitudes for the baseline have been considered. \textbf{(b)} The first two latents are fixed, and latent $2$ is sampled twice (anchor and positive). The fine-grained control offered by the multiple variables achieves better scores \wrt the single latent baseline also when NVAE is used, demonstrating the general properties of the proposed approach.

\begin{table}[tb]
\centering
\caption{Evaluation on Cifar$10$ \citep{cifar100} of encoders trained using an NVAE \citep{nvae} with $3$ latent groups as a data source.}
\label{tab:nvae_ablation}
\begin{tabular}{l|l|c}
    \toprule
     Encoder & $T_{\mathbf{z}}$ & Top-1 Accuracy (\%) \\ 
    \hline
    Baseline real & - & $79.15$ \\
    Baseline synth & - & $54.95$ \\
    \hline
    Random & $N^t(0., 0.2, 1.)$ & $59.44$ \\
    Random & $N^t(0., 0.3, 1.)$ & $59.61$ \\
    Random & $N^t(0., 0.4, 1.)$ & $59.34$ \\
    \hline
    ML Random & $N^t(0., 0.3, 1.)$ & $59.21$ \\
    ML Random & $N^t(0., 0.4, 1.)$ & $60.75$ \\
    ML Random & $N^t(0., 0.5, 1.)$ & $59.98$ \\
    \bottomrule
\end{tabular}
\end{table}

% \begin{table}[tb]
% \centering
% \begin{tabular}{c|l|l|c|c}
%     \toprule
%      & Run & $T_{\mathbf{z}}$ & Top-1 Acc & Top-5 Acc \\ 
%     \hline
%    % a) &\text{synth} & - & 43.18 & 89.21 \\
%     % \hline
%     & random & $N^t(0., 0.05, 2.)$ & $40.20$ & $88.05$ \\
%    \textbf{(a)} & random & $N^t(0., 0.10, 2.)$ & $42.09$ & $88.88$ \\
%     & random & $N^t(0., 0.15, 2.)$ & $40.15$ & $88.19$ \\
%     \hline
%    \textbf{(b)} & ML random & code $2$ resample & $\mathbf{43.86}$ & $\mathbf{89.69}$ \\
%     \bottomrule
% \end{tabular}
% \caption{Evaluation on Cifar$10$ \citep{cifar100} of encoders trained using an NVAE \citep{nvae} with $3$ latent codes as a data source. \textbf{(a)} Positive views are obtained with \textit{random} perturbation, as in \citep{gen-rep}. \textbf{(b)} positive views are obtained by resampling and substitution of the last code only, leaving the first two variables unchanged.}
% \label{tab:nvae_ablation}
% \end{table}
    
\section{Transfer Learning}
\label{app:transfer}

\Cref{tab:bigbigan_transfer_complete} contains the results for each transfer classification learning experiment performed on top of the SimSiam pre-trained encoders using BigBiGan / ImageNet-$1$K. The results refer to the last column of \Cref{tab:res-bigbigan}, where only the mean Top-$1$ accuracy over the $7$ target datasets has been reported. For a better comparison of encoders' generalization capabilities, each run was tested with $5$ different seeds, and the mean Top-$1$ accuracy was taken.\\
\newline
On each dataset, the results are computed using the test set where available, otherwise on the validation set, maintaining the original splits. For DTD \citep{DTD} the first split between the proposed ones has been employed, while for Caltech$101$ \citep{caltech101} we selected a random split of $30$ train images per class, using the remaining for testing. In this case, all the background images (distractors) have been removed.   

% complete table for transfer learning and data preprocessing for each dataset
\begin{table*}[tb]
    \centering
    \begin{tabular}{l cccccccc}
        \toprule
        Encoder & \multicolumn{7}{c}{Top-$1$ Accuracy on Target Dataset} \\
        \cmidrule(lr){2-8} 
         & Birdsnap & Caltech$101$ & Cifar$100$ & DTD & Flowers$102$ & Food$101$ & Pets \\
        \midrule
        Baseline real & $\mathbf{63.1 \pm 0.3}$ & $\mathbf{83.1 \pm 1.0}$ & $26.2 \pm 0.7$ & \underline{$\mathbf{56.4 \pm 0.3}$} & $\mathbf{59.8 \pm 2.6}$ & \underline{$\mathbf{51.5 \pm 0.2}$} & \underline{$\mathbf{67.6 \pm 0.4}$} \\
        Baseline synth & $46.3 \pm 0.4$ & $67.9 \pm 1.6$ & $\mathbf{33.4 \pm 0.4}$ & $47.7 \pm 0.5$ & $46.7 \pm 0.9$ & $41.6 \pm 0.2$ & $47.1 \pm 1.2$ \\
        \midrule
        random & $45.4 \pm 0.3$ & $68.9 \pm 0.9$ & $31.7 \pm 0.7$ & $47.9 \pm 0.6$ & $46.4 \pm 0.8$ & $42.3 \pm 0.3$ & $46.5 \pm 1.3$ \\
        ML random & \underline{$\mathbf{64.3 \pm 0.6}$} & \underline{$\mathbf{84.4 \pm 0.3}$} & \underline{$\mathbf{41.1 \pm 0.8}$} & $\mathbf{54.9 \pm 1.0}$ & \underline{$\mathbf{63.2 \pm 0.6}$} & $\mathbf{50.1 \pm 0.3}$ & $\mathbf{59.5 \pm 0.3}$ \\
        %HL random (D) & \underline{$\mathbf{65.2 \pm 0.2}$} & \underline{$\mathbf{85.1 \pm 0.3}$} & $40.4 \pm 0.2$ & $54.5 \pm 0.4$ & \underline{$\mathbf{66.8 \pm 0.6}$} & \underline{$\mathbf{52.0 \pm 0.3}$} & $\mathbf{59.5 \pm 1.0}$ \\
        \midrule
        learned & $42.0 \pm 0.3$ & $72.9 \pm 0.8$ & $31.7 \pm 0.8$ & $48.8 \pm 0.5$ & $45.2 \pm 0.7$ & $38.6 \pm 0.4$ & $44.0 \pm 0.8$ \\
        ML learned & $\mathbf{57.7 \pm 0.5}$ & $\mathbf{77.5 \pm 0.5}$ & $\mathbf{36.7 \pm 0.4}$ & $\mathbf{53.1 \pm 0.4}$ & $\mathbf{60.3 \pm 0.6}$ & $\mathbf{47.6 \pm 0.5}$ & $\mathbf{51.8 \pm 0.2}$ \\
        %HL learned (D) & $\mathbf{60.6 \pm 0.2}$ & $\mathbf{79.5 \pm 0.4}$ & $\mathbf{36.7 \pm 0.3}$ & $\mathbf{54.4 \pm 0.8}$ & $\mathbf{62.6 \pm 0.8}$ & $\mathbf{50.2 \pm 0.5}$ & $51.0 \pm 0.6$ \\
        \bottomrule
    \end{tabular}
   \caption{Transfer classification Top-$1$ accuracy's results for each pre-trained encoder using the SimSiam framework and BigBiGan/ImageNet-$1$K as a data source. $7$ different target datasets have been tested.}
    \label{tab:bigbigan_transfer_complete}
\end{table*}
    
\section{Qualitative Visualizations}
\label{app:qualitative}

\begin{figure}
    \centering
    \includegraphics[width=0.8\linewidth]{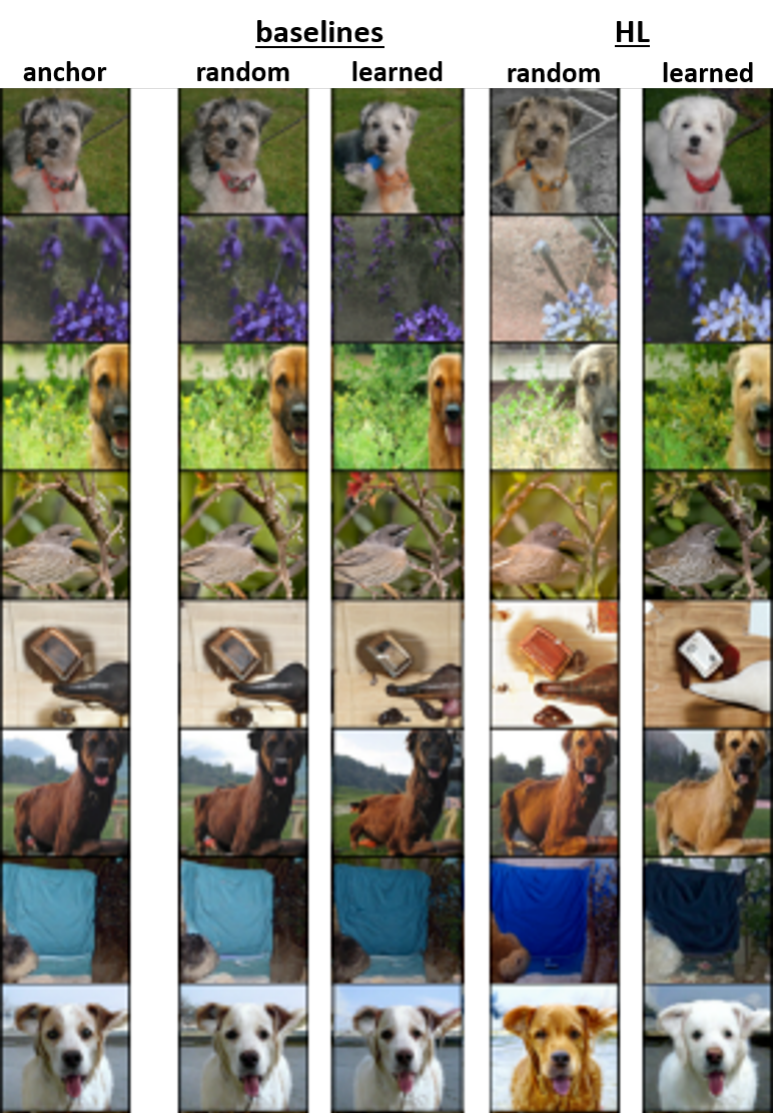}
    \caption{Example of generated views from BigBiGan anchors. From left to right: anchor, \textit{random} and \textit{learned} baseline views, \textit{random} and \textit{learned} ML views.}
    \label{fig:H_1}
\end{figure}

\begin{figure}
    \centering
    \includegraphics[width=0.8\linewidth]{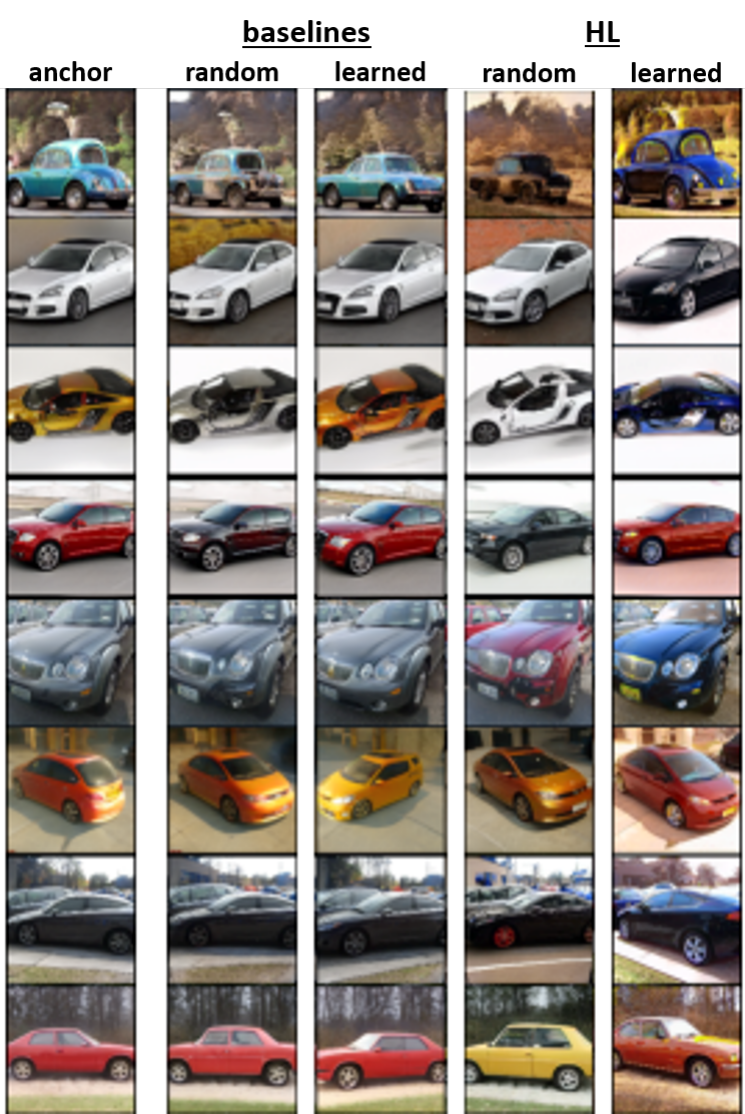}
    \caption{Example of generated views from StyleGan2 anchors. From left to right: anchor, \textit{random} and \textit{learned} baseline views, \textit{random} and \textit{learned} ML views.}
    \label{fig:H_2}
\end{figure}

\Cref{fig:H_1} and \Cref{fig:H_2} show some examples of generated views using BigBiGan and StyleGan2, respectively. In each Figure, rows represent different views, and columns display (from left to right): the initial anchor image, \textit{random} \vs \textit{learned} baselines, and the views obtained with our method. All the examples are generated using the same hyperparameters, which were also used for training the contrastive encoders. In order to allow a better comparison of the different methods, no further $T_{\mathbf{x}}$ augmentations are applied on top of the generated images. As depicted in both figures, our generated views can produce a wide range of transformations, including realistic color changes that allow us to eliminate the color jittering $T_\mathbf{x}$ augmentation.

\end{document}